\documentclass[11pt]{article}
\usepackage{siunitx}    
\usepackage{listings}
\usepackage{tabularx}
\usepackage{graphicx}
\usepackage{booktabs}
\usepackage{tcolorbox}

\usepackage{colortbl}
\usepackage{adjustbox}
\usepackage{pifont}
\newcommand{\cmark}{\textcolor{green!60!black}{\ding{51}}}
\newcommand{\xmark}{\textcolor{red!80!black}{\ding{55}}}

\usepackage{amsmath}
\usepackage{multirow}
\usepackage{makecell}
\usepackage{xcolor}
\usepackage{amssymb}
\usepackage{array}
\usepackage{algorithm}
\usepackage{algpseudocode}
\usepackage{longtable}

\usepackage[final]{acl}

\usepackage{times}
\usepackage{latexsym}
\usepackage{url}
\usepackage{hyperref}
\usepackage[T1]{fontenc}
\usepackage[utf8]{inputenc}
\usepackage{microtype}
\usepackage{inconsolata}
\usepackage{graphicx}
\usepackage{booktabs}
\usepackage{array}
\usepackage{multirow}
\usepackage{makecell}
\usepackage{geometry}

\title{Beyond Isolated Dots: Benchmarking Structured\\
Table Construction as Deep Knowledge Extraction}

\author{
\textbf{Tianyun Zhong\textsuperscript{1,2}},
\textbf{Guozhao Mo\textsuperscript{1,2}},
\textbf{Yanjiang Liu\textsuperscript{1,2}},
\textbf{Yihan Chen\textsuperscript{1,2}},
\\
\textbf{Lingdi Kong\textsuperscript{1,2}},
\textbf{Xuanang Chen\textsuperscript{2}}\thanks{Corresponding author.},
\textbf{Yaojie Lu\textsuperscript{2}},
\textbf{Hongyu Lin\textsuperscript{2}},
\\
\textbf{Shiwei Ye\textsuperscript{1}},
\textbf{Shanshan Jiang\textsuperscript{3}},
\textbf{Bin Dong\textsuperscript{3}},
\textbf{Le Sun\textsuperscript{2}},
\\
\textbf{Ben He\textsuperscript{1,2}},
\textbf{Xianpei Han\textsuperscript{2}}
\\
\textsuperscript{1}University of Chinese Academy of Sciences
\\
\textsuperscript{2}Institute of Software, Chinese Academy of Sciences
\\
\textsuperscript{3}Ricoh Software Research Center Beijing Co., Ltd.
\\
\texttt{zhongtianyun2023@iscas.ac.cn}
}

\begin{document}
\maketitle
\begin{abstract}
Large language models (LLMs) are increasingly expected to extract and synthesize structured information from complex real-world documents, yet they typically produce unstructured, hard-to-verify paragraph-style outputs. We introduce \textbf{AOE (Arranged and Organized Extraction)}\footnote{
\parbox[t]{0.9\linewidth}{
Dataset: \url{https://huggingface.co/datasets/tianyumyum/AOE}\\
Code: \url{https://github.com/tianyumyum/AOE}
}}, a bilingual benchmark that evaluates LLMs' ability to distill long-form, multi-document inputs (64k--1.8M tokens) into structured tables. AOE comprises 12 tasks across four domains---academic, financial, legal, and medical---each grounded in genuine analytical needs, with schema guidance provided to isolate extraction quality from formatting variance. We evaluate 12 frontier LLMs (both open-weight and proprietary) under four generation settings of increasing autonomy: direct full-document input, RAG-based retrieval, offline document-browsing agent, and online web-search agent. Even the best-performing model achieves only 42.7\% Cell F1, and retrieval-augmented approaches often degrade rather than improve performance, revealing that structured extraction from long contexts remains a fundamental challenge---an \textbf{``Area of Effect'' (AOE)} that impacts all tested systems. 

\end{abstract}

\section{Introduction}

\begin{figure}
    \centering
    \includegraphics[width=0.99\linewidth]{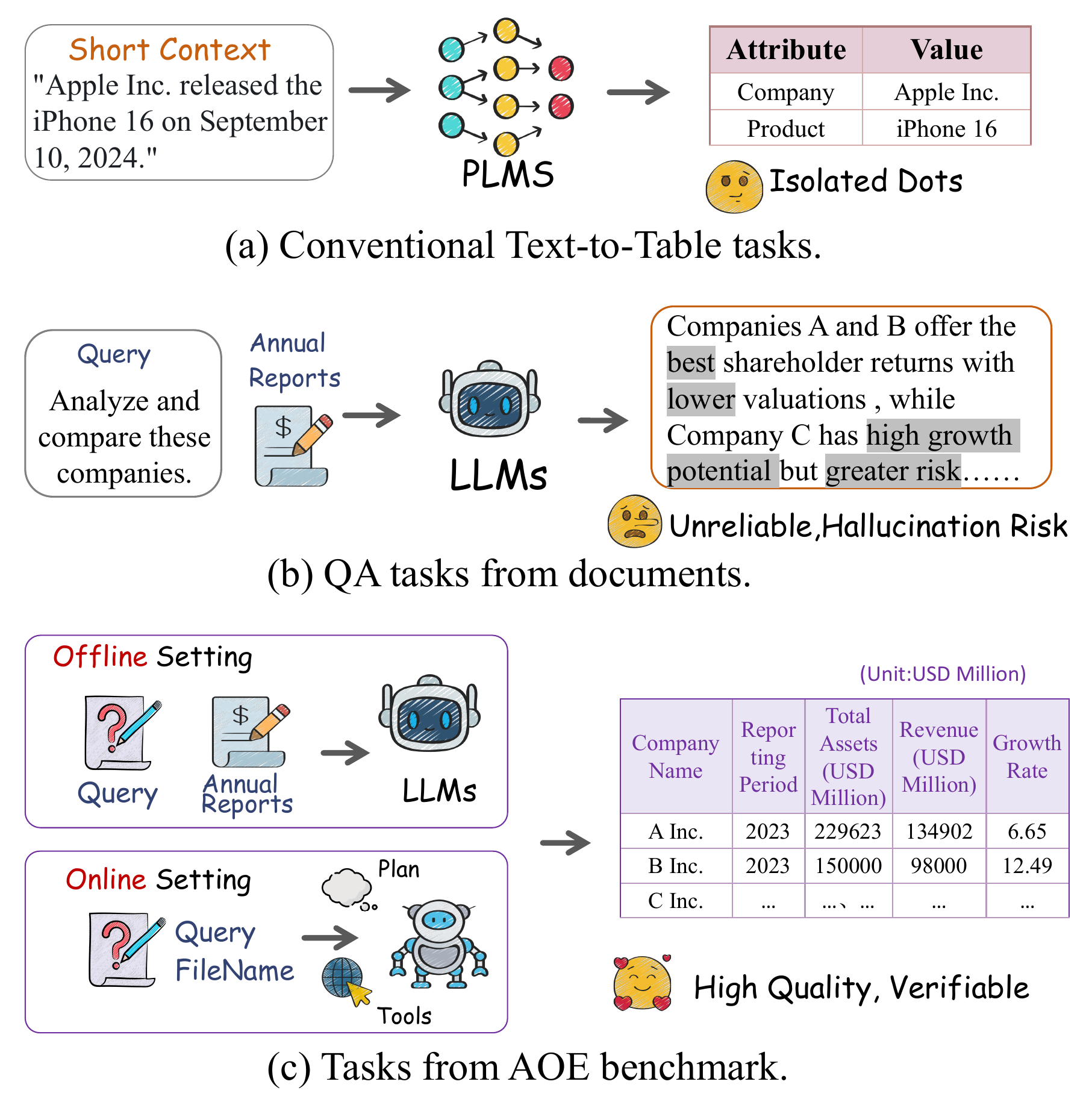}
    \caption{
    Differences between AOE and previous tasks. Unlike (a) extracting "Isolated Dots" or (b) generating "Unverifiable Output," (c) AOE requires constructing "Organized Evidence" for verifiable analysis.
    }
    \label{fig:figure_1_comparison}
\end{figure}

\begin{figure*}
    \centering
    \includegraphics[width=0.94\linewidth]{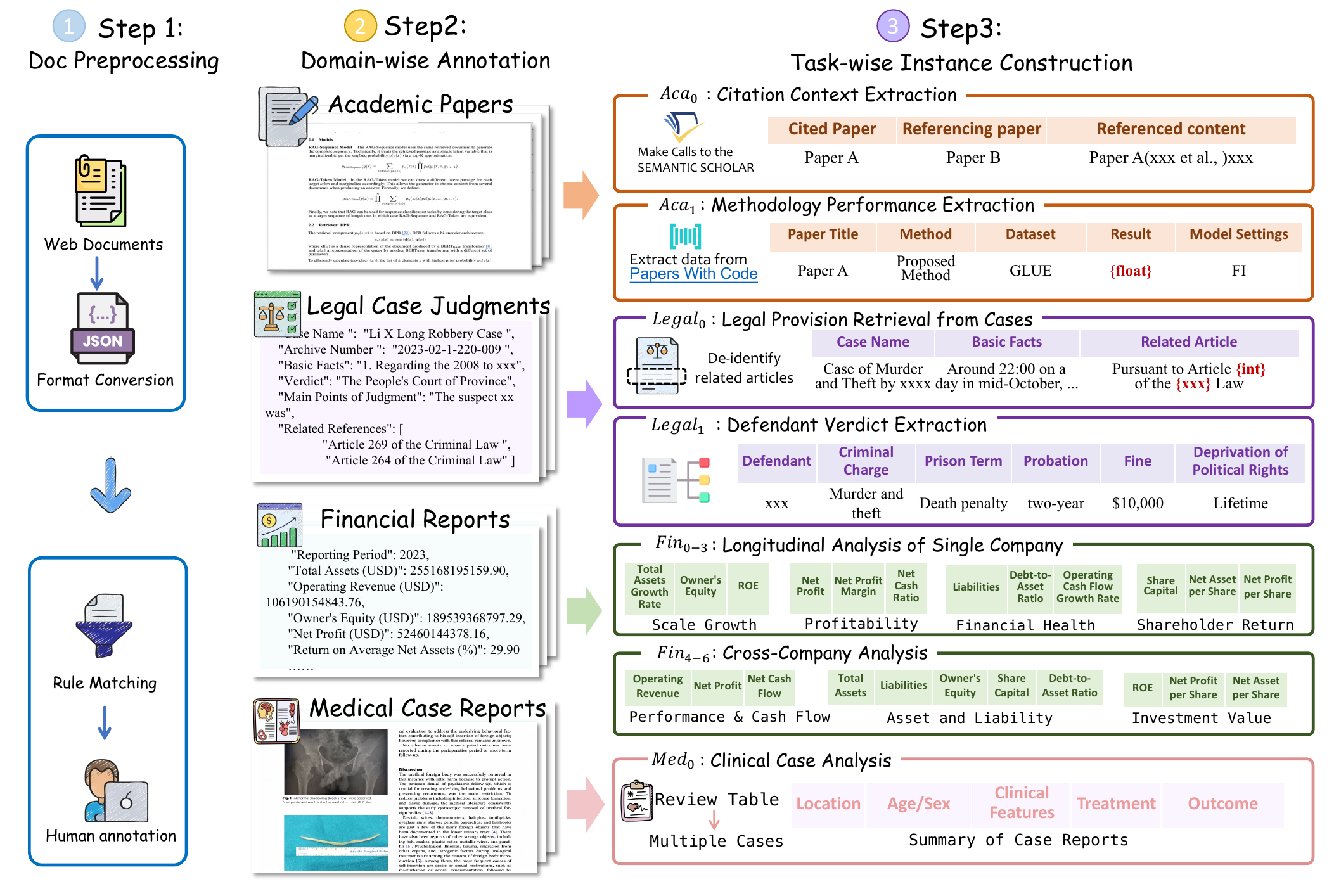}
    \caption{Construction pipeline of AOE Benchmark.
}
    \vspace{-0.2cm}
    \label{fig:construction}
\end{figure*}

Real-world analytical tasks---comparing financial performance across annual reports, summarizing clinical evidence across case studies, aggregating experimental results across papers---demand that information scattered in multiple long documents be distilled into structured, verifiable tables.
Large language models (LLMs) are increasingly expected to serve as such automated analysts~\citep{Loong,16bai2023longbench,zhangmarathon,li2025structrag}, yet they often produce paragraph-style answers that are difficult to verify and compare (Figure~\ref{fig:figure_1_comparison}(b)).

No existing benchmark adequately evaluates this structured synthesis capability. Long-context benchmarks~\citep{Loong, Leval} test comprehension without requiring structured outputs; text-to-table benchmarks~\citep{DocTabQA, ttt, TKGT} require structure but are limited to short, single-document inputs; context learning benchmarks~\citep{clbench, clbenchlife} operate on single documents. None jointly evaluates multi-document reasoning and structured knowledge construction over long contexts---precisely the capability that real-world analysis demands.

To bridge this gap, we introduce \textbf{AOE (Arranged and Organized Extraction)}, a bilingual benchmark spanning four domains (academic, financial, legal, and medical) with 2,166 source documents and an average input length of 171k tokens (Figure~\ref{fig:figure_1_comparison}(c)). AOE is the first benchmark to jointly evaluate long-context multi-document comprehension and structured table construction, with three key design principles:

\paragraph{Naturally Grounded Tasks:} Every task originates from a genuine analytical need, requiring synthesis of up to 13 documents (64k--1.8M tokens) into structured tables.
\paragraph{Hierarchical Evaluation:} An automated pipeline assesses structural parsability, overall quality via LLM-evaluator, and cell-level factual accuracy (Cell F1).
\paragraph{Multiple Generation Settings:} Beyond direct full-document input, we evaluate RAG, offline agent, and online agent settings to probe different capability axes.

Even the best-performing model achieves only 42.7\% \textit{Cell F1}, leaving over half of target facts missed or incorrect and demonstrating a widespread \textbf{``Area of Effect'' (AOE)} impact across all tested systems. Our contributions are:

\begin{itemize}
    \item We introduce AOE, the first benchmark that jointly evaluates long-context multi-document comprehension and structured table construction, spanning four domains with 2,166 source documents.
    \item We design a hierarchical evaluation framework combining structural validation, LLM-based quality scoring, and cell-level factual F1, with verified cross-judge consistency and run-to-run stability.
    \item We conduct a comprehensive study of twelve frontier LLMs across four generation settings (Direct, RAG, Offline Agent, Online Agent), revealing that current models and retrieval-augmented pipelines remain far from solving real-world structured extraction.
\end{itemize}

\section{AOE Benchmark}

Our benchmark evaluates LLMs on multi-step extraction and inferential reasoning tasks involving implicit calculations across Academic, Financial, Legal, and Medical domains, with statistical summaries and task definitions detailed in Table~\ref{tab:stats_benchmark} and Table~\ref{tab:tasks_definition} respectively.

\newcommand{\goodmark}{\ding{51}}
\newcommand{\badmark}{\ding{55}}

\begin{table*}[htbp]
  \centering
  \setlength{\tabcolsep}{2.5pt}
  \adjustbox{max width=\textwidth}{
  \begin{tabular}{l|cccccccc}
    \toprule
    \textbf{Benchmark} & \textbf{Languages} & \textbf{Task Type} & \textbf{Domain} & \textbf{\#Docs} & \textbf{Instances} & \textbf{Multi-Doc} & \textbf{Context} & \textbf{Structured Output} \\
    \midrule
    \rowcolor{gray!20}
    \multicolumn{9}{c}{\textit{Long-Context Understanding \& Context Learning}} \\
    \midrule
    L-EVAL~\cite{Leval} & EN, ZH & Long-Context Understanding & 19 & 508 & 2,000+ & \xmark & 3k--200k & \xmark \\
    Loong~\cite{Loong} & EN, ZH & Multi-Doc Long-Context QA & 3 & 1,600 & 1600 & \cmark & 110K & \xmark \\
    LongBench v2~\cite{bai2025longbench} & EN & Long-Context Multitask QA & $\sim$15+ & -- & 503 & \cmark & 104k & \xmark \\
    CL-bench~\cite{clbench} & EN & Context Learning & 4 & 500 & 1,899 & \xmark & 10.4k & \xmark \\
    
    \midrule
    \rowcolor{gray!20}
    \multicolumn{9}{c}{\textit{Text-to-Table \& Structured Construction}} \\
    \midrule
    LiveSum~\cite{ttt} & EN & Summarization to Tables & 1 & 1,122 & 1,122 & \xmark & 1.3k & \cmark \\
    CPL~\cite{TKGT} & ZH & Legal Text-to-Table w/ KG-Aug & 1 & 200 & 200 & \xmark & 1.1k & \cmark \\
    DocTabQA~\cite{DocTabQA} & EN & Long-Doc to Table & 1 & 828 & 828 & \xmark & 19.4k & \cmark \\
    \midrule
    \textbf{AOE (Ours)} & \textbf{EN, ZH} & \textbf{Multi-Doc Struct KC} & \textbf{4} & \textbf{2,166} & \textbf{447} & \cmark & \textbf{171k} & \textbf{\cmark} \\
    \bottomrule
  \end{tabular}
  }
   \caption{Comparison of AOE with existing benchmarks in Long-Context Understanding and Structured Construction. KC: Knowledge Construction. AOE uniquely combines long multi-document context with structured output generation across 4 domains.}
   \label{tab:full_comparison}
\end{table*}

\begin{table}[t]
\centering
\small
\setlength{\tabcolsep}{3pt}
\resizebox{\columnwidth}{!}{
\begin{tabular}{lcccc}
\toprule
\textbf{Metric} & \textbf{Academic} & \textbf{Financial} & \textbf{Legal} & \textbf{Medical} \\
\midrule
\rowcolor{gray!20}
\multicolumn{5}{c}{\textit{Basic Info}} \\
Language & EN & ZH, EN & ZH & EN \\
Tables (\#) & 74 & 224 & 75 & 74 \\
\midrule
\rowcolor{gray!20}
\multicolumn{5}{c}{\textit{Document-Level}} \\
Documents (\#) & 257 & 944 & 625 & 340 \\
Avg Tokens & 72k & 435k & 8k & 23k \\
\midrule
\rowcolor{gray!20}
\multicolumn{5}{c}{\textit{Table-Level}} \\
Rows (avg/max) & 4.0/15 & 4.2/5 & 11.6/55 & 4.6/5 \\
Columns (avg/max) & 4.0/4 & 7.4/9 & 6.3/10 & 10.9/17 \\
Docs (avg/max) & 3.5/5 & 4.2/5 & 8.3/10 & 4.6/5 \\
Cells (avg/max) & 15.9/60 & 30.8/36 & 78.5/550 & 50.6/85 \\
Total Input (chars) & 250K & 1.8M & 64K & 104K \\
\bottomrule
\end{tabular}
}
\caption{Data statistics of AOE benchmark.}
\label{tab:stats_benchmark}
\end{table}


Existing benchmarks for evaluating LLMs' structured information extraction exhibit critical limitations. Long-context QA benchmarks~\citep{Loong, Leval} focus on comprehension without structured outputs, while Text-to-table tasks are constrained by single-document inputs~\citep{DocTabQA, ttt, TKGT}, fixed schemas~\citep{ttt, TKGT}, or synthetic data~\citep{text2table, 18hsieh2024ruler, NIAH}. Agent-focused benchmarks like GAIA~\citep{gaia} and FRAMES~\citep{frames} support multi-document reasoning but lack evaluation of structured knowledge construction.

AOE spans four domains with widely varying input scales (Table~\ref{tab:stats_benchmark}): Financial tasks average 1.8M tokens, Academic 250K, Medical 104K, and Legal 64K. We further split the Financial domain into \textbf{Fin-S} ($\leq$200k tokens, 101 instances) and \textbf{Fin-L} ($>$200k tokens, 123 instances) to isolate the effect of context length on extraction performance. Fin-L instances, which often involve multi-year cross-company comparisons, frequently exceed the context windows of current models. However, key financial figures typically appear multiple times across different sections of the same report (e.g., in both the financial highlights and the detailed statements), providing redundant evidence that models can potentially exploit even under truncation. For models with bounded context windows, inputs exceeding the maximum length are truncated.

\subsection{Data Collection}
We curated the AOE dataset exclusively from publicly available, real-world long-form documents across four domains. Critically, unlike benchmarks that artificially concatenate unrelated documents to create long contexts, every AOE instance reflects a genuine multi-document analytical need: ground-truth tables are derived from real-world artifacts such as academic leaderboards, financial comparative reports, legal verdict summaries, and medical literature review tables. This ensures that the benchmark simultaneously tests practical utility and core model capabilities---long-context comprehension, cross-document reasoning, and structured output generation.

\textbf{Academic:} English research papers sourced from Semantic Scholar and Papers With Code. Ground-truth tables correspond to dataset leaderboards that aggregate experimental results across multiple papers.

\textbf{Financial:} Bilingual (Chinese and English) A-share company annual reports (2020--2023) sourced from CNINFO. Tasks mirror real analyst workflows requiring longitudinal and cross-company comparisons.

\textbf{Legal:} Chinese civil law judgments and statutory provisions collected from the People's Court Case Library. Tables capture structured verdict summaries and statutory provision mappings across related cases.

\textbf{Medical:} English biomedical case report papers from PubMed Central (PMC) Open Access Subset\footnote{https://pmc.ncbi.nlm.nih.gov/}. We search for systematic reviews in Q1/Q2 journals that contain ``summary of reported cases'' tables---structured tables where each row cites one independent case report and columns represent clinical attributes. These review tables serve as authoritative ground truth (curated by domain-expert authors and peer-reviewed). We then retrieve the original Open Access case report for each cited row as input documents. Table~\ref{tab:medical_pipeline} shows a concrete example; full details are in Appendix~\ref{app:data_collection}.

All documents were pre-processed from PDF to markdown.

\begin{table}[t]
\centering
\footnotesize
\setlength{\tabcolsep}{2pt}
\renewcommand{\arraystretch}{1.05}
\begin{tabular}{@{}p{0.95\columnwidth}@{}}
\toprule
\textbf{Example: Constructing a Medical Instance} \\
\midrule
\textbf{1. Review paper discovered via PubMed:} \citet{li2024potential} contains Table~1: summary of cases of NSCLC-to-SCLC transformation (2017--2023). \\[2pt]
\textbf{2. Ground-truth table extracted} (11 columns): \\
\quad \textit{source\_doc, Age, Sex, Country, Smoking\_status,} \\
\quad \textit{Mutational\_status, Medication\_before, Medication\_after,} \\
\quad \textit{CNS\_metastasis, OS\_after\_transformation, OS} \\
\quad Each row cites one independent case report; values curated by the review authors. \\[2pt]
\textbf{3. Source papers collected:} For each row, retrieve the cited OA case report. Filter: must be Open Access and describe exactly one case. Result: 7 of 12 rows pass $\to$ 7 input documents. \\[2pt]
\textbf{4. Human verification:} Confirm cell values are traceable to source papers, no duplicates, columns consistent. \\
\bottomrule
\end{tabular}
\caption{Medical data construction pipeline (real instance from our benchmark). The ground-truth table originates from a peer-reviewed systematic review in a Q1 journal, ensuring annotation quality without additional crowd-sourcing.}
\label{tab:medical_pipeline}
\end{table}

\begin{figure*}
    \centering
    \includegraphics[width=1\linewidth]{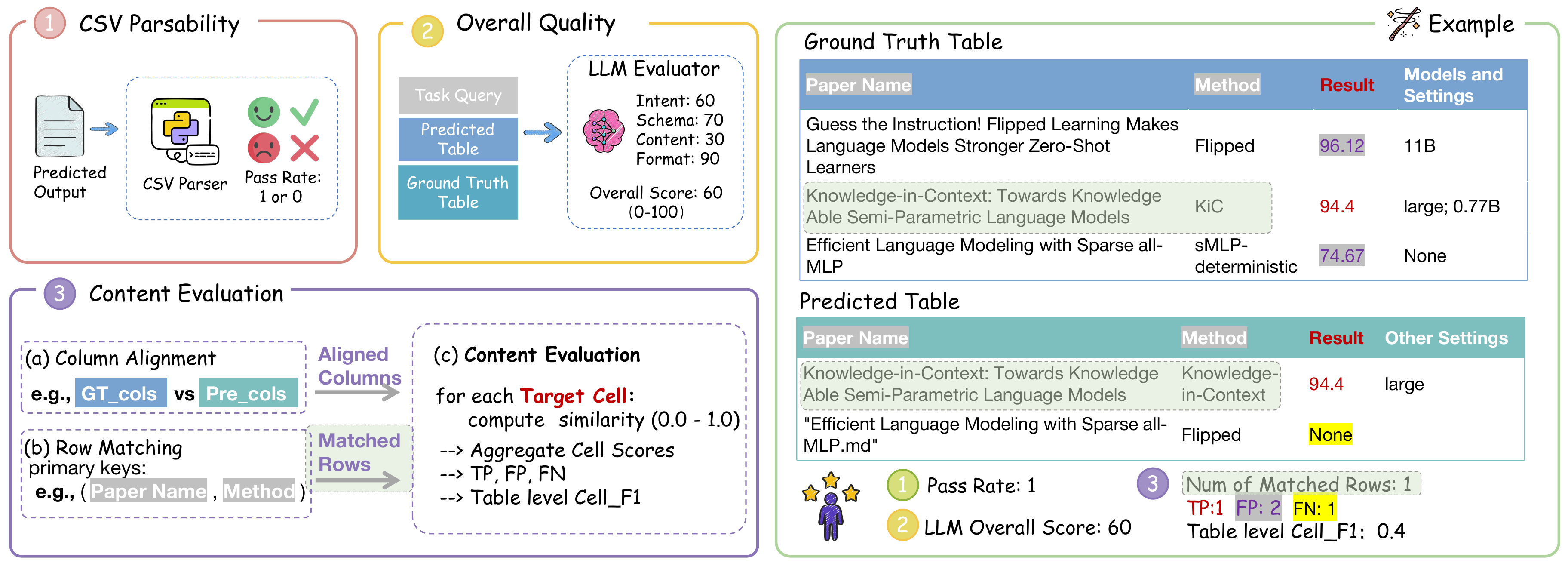}
    \caption{Overview of the Automated Evaluation Pipeline. The pipeline evaluates generated tables from three perspectives: (1) CSV Parsability for basic structural correctness; (2) Overall Quality via an LLM evaluator for a holistic score; and (3) Content Evaluation to calculate a Cell F1 score based on cell-level comparison after column and row alignment.}
    \label{fig:eval_process}
\end{figure*}

\subsection{Task Formulation}
\label{sec:task_formulation}
AOE defines 12 tasks across four domains, each requiring models to construct structured tables from multiple unstructured documents. All tasks share three core challenges:

\paragraph{Cross-Document Synthesis.} Each task instance takes multiple source documents $D=\{d_1, \dots, d_M\}$ as input (up to 13 per instance). Models must locate, align, and integrate information scattered across independent documents with heterogeneous formats and terminologies.

\paragraph{Task-Dependent Schema.} Each task defines a schema that is tightly coupled with the analytical objective of the query---columns directly reflect the real information needs (e.g., financial growth indicators, legal verdict components, clinical attributes). To ensure fair evaluation that isolates extraction quality from schema formatting, we provide explicit schema guidance in the evaluation prompt. Models must still interpret the query semantics to populate the table correctly, but are not expected to freely invent column structures.

\paragraph{Multi-Step Reasoning.} Beyond surface-level extraction, tasks require implicit calculations (e.g., year-over-year growth rates in financial reports), inferential reasoning (e.g., mapping case facts to statutory provisions), and cross-reference resolution (e.g., aligning clinical data reported in different formats across papers).

The four domains---Academic, Financial, Legal, and Medical---are designed to cover a broad spectrum of reasoning demands, including citation analysis, longitudinal and cross-entity financial comparison, legal provision retrieval and verdict extraction, and clinical evidence synthesis. Detailed per-task definitions are provided in Appendix~\ref{sec:task_details}.

\subsection{Ground-Truth Annotation Process}
\label{sec:annotation}
To ensure reproducibility and high quality, we implemented a rigorous two-phase annotation process involving domain experts.

\paragraph{Phase 1: Expert-Led Schema Design.}
This phase was led by domain experts (practitioners in finance and law). Experts defined meaningful queries simulating practical analytical needs and constructed structured table schemas (defining columns and data types) grounded in the query's analytical objective. The resulting schema serves as the gold standard for evaluation. At inference time, models receive the query along with schema guidance specifying the expected columns, ensuring that evaluation focuses on extraction and reasoning quality rather than schema formatting (see Section~\ref{sec:task_formulation}).

\paragraph{Phase 2: Independent Annotation and Adjudication.}
The core annotation team consisted of graduate students with relevant domain expertise. Each task instance was assigned to two annotators who worked independently. The resulting tables were automatically compared, and discrepancies were escalated to senior domain experts for final adjudication.
Annotators do not read documents linearly but leverage search and navigation tools to locate relevant information efficiently, completing tables in approximately 15--25 minutes each. Concrete examples of annotation guidelines and adjudicated tables are provided in Appendix~\ref{sec:appendix_annotation}.

\paragraph{Annotation Quality.}
To verify data quality, we conducted a blind re-annotation study on 436 cells across all four domains. The initial Exact Match rate was 98\%, and after expert adjudication of discrepancies, the final agreement reached 100\%, confirming that the benchmark consists of objective, verifiable facts. 

\subsection{Benchmark Evaluation}

\subsubsection{Task Definition}

Given a set of source documents $D=\{d_1, \dots, d_M\}$ and a natural language query $q$ describing the analytical goal, the task is to produce a structured table $T = (C, R)$, where $C = \{c_1, \dots, c_K\}$ is the set of columns defined by the task schema, and $R = \{r_1, \dots, r_N\}$ is a set of rows, with each cell $v_{ij}$ containing information extracted or inferred from $D$. The schema $C$ is provided to the model as part of the evaluation prompt (schema guidance); the number of rows $N$ must be determined from the document content.

\subsubsection{Automated Evaluation Pipeline and Metrics}

AOE is designed with a hierarchical evaluation pipeline that examines generated tables from basic parsability to fine-grained cell-level accuracy. We report three primary metrics: (1) \textit{LLM Score (Overall Quality)}, (2) \textit{Row F1 (Row Retrieval)}, and (3) \textit{Cell F1 (Content Evaluation)}. The pipeline is illustrated in Figure~\ref{fig:eval_process}.

The process begins by mapping model-generated tables to the ground-truth schema—a necessary preprocessing step to enable fine-grained evaluation. We then evaluate table quality using the three core metrics described below (see Appendix~\ref{app:prompt_details} for all prompt details).

\paragraph{LLM Score (Overall Quality)}  

We prompt a strong LLM evaluator to rate the overall table on a 0--100 scale across four dimensions: Intent Understanding, Schema Construction, Content Accuracy, and Format Compliance. The final score is computed as a weighted average of these dimensions with weights of 0.2, 0.3, 0.3, and 0.2, respectively.

\paragraph{Row F1 (Row Retrieval)}  
After aligning columns via an LLM-based mapper (yielding matched columns $C_{\text{aligned}}$.), we identify correctly predicted rows by matching against ground-truth primary keys. 
We match rows between predicted and ground truth tables using pre-defined primary keys, obtaining a set of matched rows $R_{\text{matched}}$. Row F1 is computed as the harmonic mean of row-level precision and recall.

\begin{table*}[t]
\centering
\small
\setlength{\tabcolsep}{4.5pt}
\renewcommand{\arraystretch}{1.08}

\begin{tabular}{@{}lcc ccccc|cccc@{}}
\toprule
\multirow{2}{*}{\textbf{Model}}
& \multirow{2}{*}{\makecell[c]{\textbf{Ctx.}\\\textbf{Limit}}}
& \multirow{2}{*}{\makecell[c]{\textbf{Pass}\\\textbf{Rate}}}
& \multicolumn{5}{c|}{\textbf{Cell F1 by Domain}}
& \multicolumn{4}{c}{\textbf{Overall Evaluation}} \\
\cmidrule(lr){4-8}
\cmidrule(lr){9-12}
& & &
\textbf{Acad.}
& \textbf{Legal}
& \textbf{Fin-S}
& \textbf{Fin-L}
& \textbf{Med.}
& \textbf{LLM}
& \textbf{Col.}
& \textbf{Row}
& \textbf{Cell} \\

\midrule
\multicolumn{12}{@{}l}{\textit{Open-Weight Models}} \\[1pt]

DeepSeek-V3.2
& 131k & 92.2
& 36.9 & 45.5 & 57.1 & 30.1 & 21.6
& 74.1 & 73.7 & 76.0 & 38.2 \\

Qwen-3.6-Plus
& 131k & 94.4
& 52.2 & 48.2 & 48.8 & 22.7 & 26.0
& \textbf{75.5} & 74.3 & 78.2 & 39.6 \\

Kimi-K2.6
& 131k & 96.0
& 34.5 & 43.3 & \textbf{63.4} & \textbf{35.3} & 23.4
& 73.8 & 82.8 & 75.2 & 40.0 \\

GLM-5.1
& 200k & 96.2
& 42.1 & 44.4 & 52.9 & 33.4 & \textbf{28.9}
& 74.9 & 78.0 & 78.2 & 40.3 \\

Kimi-K2.5
& 131k & \textbf{98.0}
& 34.1 & 50.0 & 62.5 & 35.0 & 28.4
& 74.0 & \textbf{83.0} & 76.7 & \textbf{42.0} \\

DeepSeek-V4-Pro
& 131k & 96.2
& \textbf{58.6} & \textbf{50.4} & 52.8 & 24.6 & 23.7
& 72.9 & 76.1 & \textbf{82.3} & \textbf{42.0} \\

\midrule
\multicolumn{12}{@{}l}{\textit{Proprietary Models}} \\[1pt]

Gemini-3.1-Pro
& 131k & 87.9
& 42.8 & \textbf{44.3} & 53.1 & 27.1 & 16.1
& 68.7 & 68.0 & 71.6 & 36.7 \\

Gemini-3-Flash
& 1M & 77.4
& 44.2 & 38.3 & 44.4 & \textbf{55.1} & 6.1
& 69.6 & 64.3 & 61.9 & 37.6 \\

Claude-Sonnet-4.6
& 200k & 98.7
& 40.2 & 36.6 & \textbf{60.4} & 32.7 & 27.6
& 75.3 & \textbf{86.2} & 73.4 & 39.5 \\

GPT-5.4
& 128k & 94.9
& \textbf{48.2} & 40.1 & 59.9 & 32.7 & 18.7
& 75.3 & 80.5 & 76.9 & 39.9 \\

Claude-Opus-4.6
& 200k & \textbf{99.6}
& 36.3 & 39.0 & 56.0 & 38.6 & \textbf{31.4}
& \textbf{75.4} & 83.7 & \textbf{80.4} & 40.3 \\

Seed-2.0-Pro
& 262k & 93.5
& 45.3 & 40.0 & 60.1 & 40.7 & 25.6
& 75.1 & 80.2 & 79.1 & \textbf{42.7} \\

\bottomrule
\end{tabular}

\caption{
\textbf{Main results on AOE.}
We report domain-level Cell F1 and overall evaluation scores:
a holistic LLM-as-judge score (\textbf{LLM}),
Column F1 (\textbf{Col.}), Row F1 (\textbf{Row}), and
Cell F1 (\textbf{Cell}).
Financial instances are split into \textbf{Fin-S}
($\leq$200k total input tokens) and \textbf{Fin-L}
($>$200k total input tokens).
\textbf{Ctx. Limit} denotes the maximum context length allowed
in our evaluation pipeline.
Models are sorted by overall Cell F1 in ascending order
within each category.
Bold indicates the best result within each model category
and evaluation column.
F1 scores and Pass Rate are reported in \%;
LLM scores are on a 0--100 scale.
}
\label{tab:main_results_v2}
\end{table*}

\paragraph{Cell F1 (Content Evaluation)}
\label{sec:cell_f1}
Cell F1 operates at two levels. At the \emph{cell level}, an LLM evaluator
assigns each aligned predicted--gold cell pair a semantic matching score
$s_{ij}\in[0,1]$. This allows cells containing complex or multi-item content
to receive partial credit rather than requiring exact string matching.
For list or array values, partial matches are scored according to item-level
coverage with a penalty factor of $0.8$ for incomplete or extraneous content
(see Appendix~\ref{app:cell_rating} for the detailed scoring rules).



At the table level, we aggregate the cell-level semantic
matching scores into precision, recall, and F1.
Let $\mathcal{A}$ denote the set of aligned predicted--gold
cell pairs, with matching scores $s_{ij}\in[0,1]$.
Let $N_{\mathrm{pred}}$ and $N_{\mathrm{gold}}$ denote the
total numbers of cells in the predicted and gold tables,
respectively. We compute
\begin{align}
S &= \sum_{(i,j)\in\mathcal{A}} s_{ij}, \\
P_{\mathrm{cell}} &= \frac{S}{N_{\mathrm{pred}}},
\qquad
R_{\mathrm{cell}} = \frac{S}{N_{\mathrm{gold}}}, \\
F1_{\mathrm{cell}} &=
\frac{2P_{\mathrm{cell}}R_{\mathrm{cell}}}
{P_{\mathrm{cell}}+R_{\mathrm{cell}}}.
\end{align}
Unmatched cells contribute zero to $S$ but remain in
their respective denominators, while partially matched
cells contribute their semantic matching scores.
Given fixed alignments and cell-level scores,
table-level aggregation is deterministic.

Thus, the LLM evaluator is used only for semantic matching at the individual
cell level, whereas table-level aggregation follows the standard F1
formulation.

\paragraph{Evaluator Robustness.}
To verify the reliability of our LLM-based scoring pipeline, we conducted stability analyses across three dimensions: (1) \textit{Inter-judge consistency}: two independent LLM judges evaluated 6,866 cells, yielding a Pearson correlation of 0.9134 ($p < 0.001$); (2) \textit{Intra-judge consistency}: repeated scoring (8 runs) on 5,636 cells showed variance $\leq 0.03$ across all value types; (3) \textit{Ranking robustness}: bootstrap analysis confirmed model rankings shift by less than 1 position on average. Full details are provided in Appendix~\ref{app:multi_run}.

\newcommand{\taskdetailsplaceholder}{}
\label{sec:task_details_placeholder}
\section{Experiments}
\label{sec:experiments}

\subsection{Experimental Setup}
\label{subsec:setup}


\paragraph{Models Under Evaluation.}
We evaluate twelve frontier LLMs. \textbf{Open-weight models} include DeepSeek-V3.2~\citep{deepseekv3}, DeepSeek-V4-Pro~\citep{deepseekv4}, Qwen-3.6-Plus~\citep{qwen3_6}, GLM-5.1~\citep{GLM-5_1}, and Kimi-K2.5/K2.6~\citep{kimi_k2_5,kimi_k2_6}. \textbf{Proprietary models} include GPT-5.4~\citep{gpt5_4}, Claude-Opus-4.6 and Claude-Sonnet-4.6~\citep{claude}, Gemini-3.1-Pro and Gemini-3-Flash~\citep{gemini3}, and Seed-2.0-Pro~\citep{seed2}. We use GPT-4o~\citep{gpt4_o} as the LLM judge for cell-level scoring.

\paragraph{Generation Settings.}
Our primary evaluation uses the \textbf{Direct} setting, where full source documents are provided for single-pass inference. We additionally evaluate \textbf{RAG} (chunk-based retrieval), \textbf{Offline Agent} (iterative local document navigation), and \textbf{Online Agent} (open-web search) to study different information-access strategies. We compare these settings in Section~\ref{subsec:system_variant}, with implementation details in Appendix~\ref{app:generation_settings}.

\begin{figure*}[t]
    \centering
    \includegraphics[width=\textwidth]{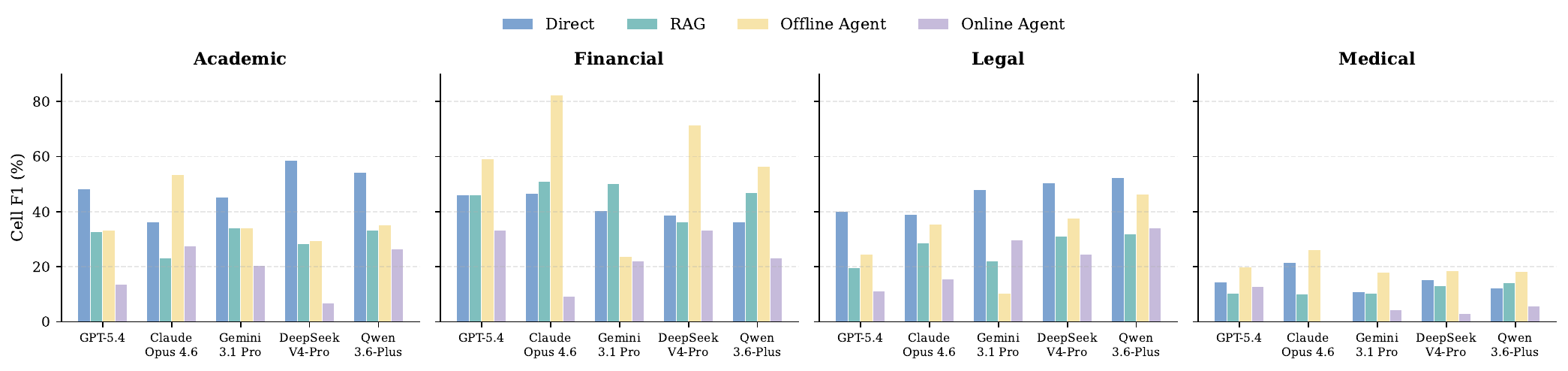}
    \caption{
    Cell F1 (\%) across four generation settings---Direct, RAG,
    Offline Agent, and Online Agent---for five representative models,
    broken down by domain.
    }
    \label{fig:settings_comparison}
\end{figure*}

\subsection{Main Results: The Bottleneck of Long Contexts}
\label{subsec:main_results}

\begin{table}[t]
    \centering
    \footnotesize
    \setlength{\tabcolsep}{1.5pt}
    \begin{tabular}{@{}ll rrrrr@{}}
        \toprule
        \textbf{Model} & \textbf{Setting} & \textbf{CF1} & \textbf{ColR} & \textbf{ColP} & \textbf{RowR} & \textbf{CSV} \\
        \midrule
        \multirow{3}{*}{\makecell[l]{Claude-\\Sonnet-4.6}}
        & Prompt-All & 19.6 & 71.5 & 63.9 & 92.6 & 64.9 \\
        & Schema-Guided & 28.4 & 93.1 & 83.9 & 92.1 & 95.9 \\
        & \cellcolor{green!8}$\Delta$ & \cellcolor{green!8}\textbf{+8.8} & \cellcolor{green!8}\textbf{+21.6} & \cellcolor{green!8}+20.0 & \cellcolor{green!8}$-$0.5 & \cellcolor{green!8}+31.0 \\
        \midrule
        \multirow{3}{*}{\makecell[l]{DeepSeek-\\V3.2}}
        & Prompt-All & 20.3 & 66.9 & 59.2 & 95.4 & 52.7 \\
        & Schema-Guided & 23.4 & 89.1 & 80.1 & 89.8 & 90.5 \\
        & \cellcolor{green!8}$\Delta$ & \cellcolor{green!8}\textbf{+3.1} & \cellcolor{green!8}\textbf{+22.2} & \cellcolor{green!8}+20.9 & \cellcolor{green!8}$-$5.6 & \cellcolor{green!8}+37.8 \\
        \midrule
        \multirow{3}{*}{\makecell[l]{Qwen-\\3.6-Plus}}
        & Prompt-All & 17.2 & 60.3 & 53.9 & 86.9 & 67.6 \\
        & Schema-Guided & 27.7 & 90.7 & 81.6 & 91.9 & 93.2 \\
        & \cellcolor{green!8}$\Delta$ & \cellcolor{green!8}\textbf{+10.5} & \cellcolor{green!8}\textbf{+30.4} & \cellcolor{green!8}+27.7 & \cellcolor{green!8}+5.0 & \cellcolor{green!8}+25.6 \\
        \bottomrule
    \end{tabular}
    \caption{Schema-Guided vs.\ Prompt-All on the Medical domain (74 instances). CF1 = Cell F1, ColR = Column Recall, ColP = Column Precision, RowR = Row Recall, CSV = CSV parse rate. All values in \%.}
    \label{tab:schema_guided_comparison}
\end{table}

Table~\ref{tab:main_results_v2} shows that AOE remains challenging for all
evaluated models: the best performer, Seed-2.0-Pro, achieves only 42.7\%
average Cell F1, leaving substantial room for improvement.

\paragraph{Similar overall scores mask heterogeneous capability profiles.}
Despite the narrow range of average Cell F1 scores (36.7--42.7\%),
performance varies substantially across domains. For example,
DeepSeek-V4-Pro achieves the highest Academic Cell F1 (58.6\%) but only
24.6\% on Fin-L, whereas Gemini-3-Flash reaches 55.1\% on Fin-L but
44.2\% on Academic. These results suggest that aggregate scores obscure
substantial differences in domain-specific capabilities.

\paragraph{Long financial inputs remain particularly challenging.}
For 11 of the 12 evaluated models, Cell F1 decreases from Fin-S to Fin-L,
often by more than 20 points. For example, DeepSeek-V4-Pro drops from
52.8\% to 24.6\%, and Claude-Sonnet-4.6 from 60.4\% to 32.7\%.
Gemini-3-Flash is the only exception, improving from 44.4\% to 55.1\%.
This pronounced degradation motivates our subsequent analysis of retrieval
and agentic approaches for handling long inputs.

\paragraph{Open-weight and proprietary models perform comparably overall.}
The strongest models from the two groups achieve similar average Cell F1
scores (42.0\% for DeepSeek-V4-Pro and 42.7\% for Seed-2.0-Pro), with
substantial overlap across the leaderboard. Model access type alone therefore
does not explain performance differences on AOE.


\subsection{Effect of Schema Guidance}
\label{subsec:schema_guidance}

A key design choice in AOE is whether to provide explicit schema guidance
(column definitions) alongside the query or allow models to freely determine
the output schema. We ablate this choice on the Medical domain, where schema
divergence is most pronounced, comparing \textbf{Prompt-All} (free-form
schema) with \textbf{Schema-Guided} (explicit column definitions).

Table~\ref{tab:schema_guided_comparison} shows that schema guidance
substantially reduces schema divergence. Column Recall increases by
21--30 points across all three models, while Cell F1 improves by
3--11 points. The gains are substantially larger for column alignment than
for row recovery, suggesting that schema guidance primarily reduces
variation in schema construction and allows Cell F1 to more directly
reflect content extraction quality. A qualitative case study is provided
in Appendix~\ref{app:schema_guided}.

\begin{table*}[t]
\centering
\small
\setlength{\tabcolsep}{5.5pt}
\renewcommand{\arraystretch}{1.08}
\begin{tabular}{lccc|cccc}
\toprule
& \multicolumn{3}{c|}{\textbf{Table Construction}}
& \multicolumn{4}{c}{\textbf{Cell Correctness}} \\
\cmidrule(lr){2-4}
\cmidrule(lr){5-8}
\textbf{Model}
& \textbf{Column}
& \textbf{Row}
& \textbf{Coverage}
& \textbf{Num.}
& \textbf{Calc.}
& \textbf{Reason.}
& \textbf{Text} \\
\midrule

DeepSeek-V3.2
& 67.6 & \textbf{89.8} & 50.6
& 90.0 & 43.9 & 41.6 & 25.7 \\

Kimi-K2.6
& 75.8 & 80.5 & 54.0
& 85.0 & 46.8 & \textbf{44.4} & 26.6 \\

Claude-Sonnet-4.6
& \textbf{77.0} & 84.1 & \textbf{55.8}
& 86.2 & 45.7 & 27.1 & \textbf{30.2} \\

Gemini-3-Flash
& 69.8 & 86.9 & 55.7
& 87.5 & \textbf{58.5} & 31.7 & 19.9 \\

GPT-5.4
& 73.8 & 88.6 & 54.0
& \textbf{91.2} & 45.1 & 36.2 & 19.2 \\

Seed-2.0-Pro
& 69.9 & 85.0 & 53.9
& 83.8 & 53.6 & 43.4 & 21.8 \\

\bottomrule
\end{tabular}
\caption{
\textbf{Error decomposition in the offline setting.}
We trace model failures through three stages of table construction:
Column recovery (Column), row recovery (Row), and cell coverage (Coverage); then further report Cell F1 by reasoning type:
numeric extraction (Num.), financial calculation (Calc.),
legal reasoning (Reason.), and textual summarization (Text).
Results for all models and raw matched/gold counts are provided in Appendix~\ref{app:error_analysis}.}
\label{tab:error_analysis}
\end{table*}

\subsection{Can Retrieval and Agents Help?}
\label{subsec:system_variant}

Beyond the Direct setting, we evaluate whether alternative mechanisms for
accessing source information can improve structured table construction
(Figure~\ref{fig:settings_comparison}).
RAG uses FAISS-based chunk retrieval, Offline Agent enables iterative
navigation over local documents, and Online Agent provides external search
and webpage-reading tools.

\paragraph{Retrieval alone does not consistently improve extraction.}
RAG substantially degrades Cell F1 on Academic and Legal tasks
(e.g., DeepSeek-V4-Pro Academic: 58.6\% $\rightarrow$ 30.7\%;
Qwen-3.6-Plus Legal: 48.2\% $\rightarrow$ 30.9\%).
In contrast, RAG can benefit some Financial tasks, where source inputs are
substantially longer. These results suggest that standard semantic retrieval
can alleviate context pressure in some long-input settings but is insufficient
for preserving the distributed evidence required by many AOE tasks.
A length-stratified analysis is provided in Appendix~\ref{app:rag_length}.

\paragraph{Agentic browsing helps primarily on ultra-long inputs.}
Offline Agent substantially improves Financial performance for several
models, including Claude-Opus-4.6 (46.7\% $\rightarrow$ 60.9\%) and
DeepSeek-V4-Pro (39.9\% $\rightarrow$ 62.0\%), suggesting that selective,
iterative document navigation can help when full-document input becomes
impractical. However, these gains do not consistently transfer to Academic
and Legal tasks, where Offline Agent often underperforms Direct inference.
Online Agent performs worst overall across most model--domain combinations,
showing that greater autonomy and external information access do not
necessarily translate into more accurate structured extraction.

\subsection{Where Do Models Fail?}
\label{subsec:error_analysis}

\textbf{Models recover structure better than content.}
As shown in Table~\ref{tab:error_analysis}, models recover most target rows
(80--90\%), yet populate only 45--56\% of gold cells.
Thus, recovering the table structure does not guarantee complete content
extraction: a substantial fraction of the target table remains unanswered
even when the relevant rows are identified.

\textbf{Extraction is easier than reasoning and synthesis.}
Among populated cells, direct numeric extraction is comparatively reliable
(84--91\% among the models shown), whereas calculation and legal reasoning
are substantially harder, and textual synthesis is the most challenging
(15--30\%).
Together, these results reveal two complementary bottlenecks in AOE:
\emph{coverage}, whether models populate the required cells, and
\emph{correctness}, whether they correctly transform and synthesize the
evidence needed for those cells.
Additional breakdowns and qualitative examples are provided in
Appendix~\ref{app:error_analysis}.

\textbf{Human performance reveals different sources of difficulty.}
On a small 16-instance reference subset, a tool-assisted graduate annotator
achieves high Cell F1 on Financial (92.9) and Legal (85.8), but substantially
lower scores on Academic (42.4) and Medical (34.0).
This contrast suggests that benchmark difficulty reflects different factors
across domains.
We provide the full human reference study in
Appendix~\ref{app:human_reference}.

\section{Related Work}

\subsection{Long-Context Understanding \& Context Learning}

Existing long-context benchmarks often suffer from limitations: synthetic tasks like NIAH~\citep{19kamradt2023} and Counting Stars~\citep{song2024counting} only test surface-level retrieval, while early datasets~\citep{16bai2023longbench, loogleli2024looglelongcontextlanguagemodels} lack sufficient length. Although recent works~\citep{Leval, dong2024bamboocomprehensivebenchmarkevaluating, qiu2024clongevalchinesebenchmarkevaluating, infinite-zhang-etal-2024-bench} and flexible methods like Ruler~\citep{18hsieh2024ruler} or Babilong~\citep{kuratov2024babilong} improve comprehensiveness, they primarily focus on passive comprehension. More recently, CL-bench~\citep{clbench} and CL-bench Life~\citep{clbenchlife} evaluate \textit{context learning}---whether models can acquire new knowledge from provided context and apply it to solve tasks. Oolong~\citep{oolong} specifically targets information aggregation over long inputs up to 128K tokens, requiring models to classify and count scattered evidence. 
However, these benchmarks operate on single-document inputs and do not require structured output generation.


\subsection{Text-to-Table Generation} 

Early approaches relied on rigid schemas~\citep{OpenTE} and synthetic data~\citep{text2table, gtablestext}, failing to capture real-world complexity. LLM-based methods have shifted toward formatting~\citep{structbench}, reconstruction~\citep{singh2024tabularis, ni2023unifiedtextstructuralizationinstructiontuned}, domain-specific extraction~\citep{DocTabQA, MedT2T}, and intermediate organization~\citep{ttt, TKGT}. Notably, recent advancements like ARXIV2TABLE~\citep{arxivdigestable, wang2025llmsgeneratetabularsummaries} have made strides in tabular summarization for scientific literature. However, their primary focus remains on \textit{fuzzy retrieval} and \textit{abstractive summarization} of broad concepts, rather than precise fact extraction. Furthermore, benchmarks like CT-Eval~\citep{cttable} and Text-Tuple-Table~\citep{ttt} are largely limited to single-document inputs. 

\section{Conclusion}

In this work, we introduced the AOE benchmark to systematically evaluate the end-to-end process of structured knowledge construction from complex, real-world documents. Our comprehensive evaluation of twelve state-of-the-art LLMs across four generation settings reveals that even the best model achieves only 42.7\% Cell F1, exposing fundamental capability gaps in numerical reasoning and cross-document synthesis. Furthermore, retrieval-augmented and agentic approaches do not consistently improve over direct input, and no single model dominates all domains. AOE establishes a challenging new baseline and underscores the need for future research to build truly reliable knowledge extraction systems.

\section*{Limitations}

AOE focuses on constructing flat relational tables from multi-document evidence, enabling reproducible cell-level evaluation but not covering more complex structures such as hierarchical or nested tables. Since schema guidance is provided in some settings, AOE also does not fully evaluate open-ended schema induction. Some inputs exceed the context windows of current models, although our RAG and agentic settings provide alternative mechanisms for accessing the document collection. Finally, Cell F1 relies on an LLM evaluator for cell-level semantic matching, which may introduce evaluation variance despite the consistency observed in our robustness experiments.

\section*{Acknowledgments}

We sincerely thank the reviewers for their insightful comments and valuable suggestions. This work was supported by the National Key R\&D Program of China (2024YFC3308000), the Natural Science Foundation of China (No. 62506354, 62572456 and 62306303) and the Postdoctoral Fellowship Program of CPSF under Grant Number GZC20251041.

\bibliography{main}

\appendix
\label{sec:appendix}




\section{Ethical Considerations}

AOE is constructed from publicly available academic, financial, legal,
and medical documents. We respect the access and usage conditions of the
original data sources and do not introduce additional personally identifiable
information beyond that contained in the source documents. For potentially
sensitive legal and medical content, we retain only information necessary
for benchmark construction and evaluation. Potential risks associated with
the use of structured extraction systems in high-stakes domains are discussed
in the Limitations section.

\paragraph{License and Intended Use.}
AOE is released under the Creative Commons Attribution 4.0 International
(CC BY 4.0) license for research and evaluation of structured information
extraction systems. The original source documents are not relicensed by us
and remain subject to their respective licenses and terms of use. Users are
responsible for complying with the access, use, and redistribution conditions
of the original data sources.



\section{Compute Resources}
\label{app:compute}

All twelve models evaluated in this work are accessed exclusively through
their official APIs. Open-weight models (DeepSeek-V3.2, DeepSeek-V4-Pro,
Qwen-3.6-Plus, GLM-5.1, Kimi-K2.5, and Kimi-K2.6) have publicly available
architecture details and weights documented in their respective model cards
and technical reports. Proprietary models (GPT-5.4, Claude-Sonnet-4.6,
Claude-Opus-4.6, Gemini-3.1-Pro, Gemini-3-Flash, and Seed-2.0-Pro) do not
publicly disclose parameter counts.

\paragraph{Inference Hyperparameters.}
For the Direct and RAG settings, we use \texttt{temperature}=0.
For the Offline Agent and Online Agent settings, we use
\texttt{temperature}=0.1 and \texttt{max\_tokens}=4096 per tool-call turn,
with a maximum of 10 agent iterations.

\paragraph{Context Windows and Inference Cost.}
Table~\ref{tab:model_context_cost} reports the context window used in our
experiments and the estimated per-task inference cost for representative
models. Reporting the effective context window also makes potential
truncation effects explicit; we analyze the position of ground-truth
evidence relative to these context limits in
Appendix~\ref{app:context_truncation}.
\begin{table}[t]
\centering
\small
\setlength{\tabcolsep}{7pt}
\renewcommand{\arraystretch}{1.08}
\begin{tabular}{lrr}
\toprule
\textbf{Model} &
\textbf{Context Window} &
\textbf{Cost / Task} \\
\midrule
Claude-Opus-4.6 & 200k & \$190.8 \\
GPT-5.4         & 128k & \$74.4  \\
Gemini-3.1-Pro  & 131k & \$60.3  \\
DeepSeek-V4-Pro & 131k & \$13.2  \\
Qwen-3.6-Plus   & 131k & \$11.8  \\
\bottomrule
\end{tabular}

\caption{
Context windows and estimated per-task API inference costs for
representative models.
}
\label{tab:model_context_cost}
\end{table}
\section{Context Coverage and Truncation Analysis}
\label{app:context_truncation}

One potential confounding factor in the Direct setting is input truncation:
when the concatenated source documents exceed a model's context window,
relevant evidence may fall outside the accessible context. We therefore
quantify how much ground-truth evidence occurs at different positions in
the concatenated input.

For each ground-truth cell, we locate the first occurrence of its target
content in the concatenated source documents and measure its position in
the input. Table~\ref{tab:gt_context_coverage} reports the proportion of
ground-truth cells whose first occurrence lies beyond 32k, 128k, and 256k
tokens.

\begin{table}[t]
\centering
\small
\setlength{\tabcolsep}{5pt}
\renewcommand{\arraystretch}{1.08}
\begin{tabular}{lrrrr}
\toprule
\textbf{Domain} &
\textbf{GT Cells} &
\textbf{$>$32k} &
\textbf{$>$128k} &
\textbf{$>$256k} \\
\midrule
Academic  & 1,037  & 15.5\% & 1.7\%  & 0.0\% \\
Financial & 6,031  & 24.4\% & 19.0\% & 9.9\% \\
Legal     & 4,075  & 0.0\%  & 0.0\%  & 0.0\% \\
Medical   & 3,283  & 0.3\%  & 0.0\%  & 0.0\% \\
\midrule
\textbf{Overall} & \textbf{14,426} &
\textbf{11.4\%} & \textbf{8.1\%} & \textbf{5.6\%} \\
\bottomrule
\end{tabular}
\caption{
\textbf{Position of ground-truth evidence in the input context.}
For each ground-truth cell, we locate the first occurrence of its target
content in the concatenated source documents and report the proportion
appearing beyond different token positions. Financial tasks exhibit the
largest long-context tail.
}
\label{tab:gt_context_coverage}
\end{table}

Overall, 91.9\% of ground-truth cells have their first occurrence within
the first 128k tokens, whereas only 8.1\% first appear beyond this position.
This indicates that truncation alone cannot account for the large remaining
performance gap: the best model achieves only 42.7\% Cell F1 despite the
majority of ground-truth evidence occurring within a typical 128k context
window. The effect is highly domain-dependent. In Academic, Legal, and
Medical tasks, nearly all ground-truth evidence occurs within 128k tokens,
whereas Financial exhibits a substantial long-context tail, with 19.0\% of
ground-truth cells first appearing beyond 128k tokens and 9.9\% beyond 256k.

These results suggest that context truncation is an important factor for
the longest Financial tasks, but is unlikely to be the primary explanation
for the overall extraction gap. Even when relevant evidence is accessible
within the context window, models must still identify, integrate, and map
that evidence to the appropriate table cells.

To further reduce dependence on fixed context windows, we additionally
evaluate RAG and agentic settings
(Appendix~\ref{app:rag_analysis}), which access source information
selectively rather than relying solely on a single concatenated context.
These settings provide complementary evidence about the extent to which
retrieval and iterative document access can alleviate long-context
constraints.
\section{Generation Settings}
\label{app:generation_settings}

We evaluate four generation settings with increasing levels of autonomy. All settings share the same prompt template (query + schema guidance) and output format (CSV); they differ only in how the model accesses source documents.

\subsection{Direct (Full-Document Input)}
Source documents are concatenated and provided in full as part of the prompt context. The model performs single-pass inference over the entire input. This setting tests the model's core ability to synthesize structured tables from long-context input, with document lengths ranging from 2K to 1.8M tokens across domains.

\subsection{RAG (Chunk-Based Retrieval)}

Documents are pre-processed into overlapping text chunks
(\texttt{chunk\_size}=512, \texttt{overlap}=100) using
\texttt{RecursiveCharacterTextSplitter}.
Each chunk is embedded with BGE-M3\footnote{\url{https://huggingface.co/BAAI/bge-m3}}
~\citep{bge-m3} and indexed using
FAISS\footnote{\url{https://github.com/facebookresearch/faiss}}
~\citep{faiss}.
At inference time, the top-50 most relevant chunks are retrieved via cosine
similarity and provided as context.
The pipeline is implemented with
LangChain\footnote{\url{https://github.com/langchain-ai/langchain}}
~\citep{Chase_LangChain_2022}.

\subsection{Offline Agent (Document Browsing)}
An iterative agent that navigates local documents without seeing their full content at once. The agent is equipped with two tools:

\begin{itemize}    \item \texttt{read\_toc(doc\_name)}: Returns the table of contents of a document, listing section IDs, headings, and character counts.
    \item \texttt{read\_section(doc\_name, section\_id)}: Returns the full text of a specific section (up to 8,000 characters).
\end{itemize}

The agent iteratively calls these tools to explore document structure, read relevant sections, and accumulate enough information to produce the target CSV. A maximum of 10 iterations is allowed per task. This setting tests multi-turn reasoning and selective information retrieval from structured documents.

\subsection{Online Agent (Web Search)}
The model receives \textit{only document names} as input---no document content is provided. The agent must autonomously locate and retrieve information using two external tools:

\begin{itemize}    \item \texttt{web\_search(query)}: Performs a web search via the Serper API\footnote{\url{https://serper.dev/}} and returns the top-10 search result snippets.
    \item \texttt{read\_url(url)}: Fetches and extracts the main text content from a URL via Jina Reader\footnote{\url{https://jina.ai/reader/}}.
\end{itemize}

The agent uses Serper
for web search and Jina Reader
for webpage content extraction.

The agent iterates through up to 10 search-and-read cycles
to gather sufficient information before generating the final CSV.
This setting evaluates end-to-end information seeking and
structuring when source documents are not locally available.

\section{Additional Analysis of Generation Settings}
\label{app:generation_analysis}

\subsection{Performance by Input Length}
\label{app:rag_length}

To better understand when retrieval-based methods are beneficial, we
stratify instances by source-document length and compare Direct and RAG
performance within each group.

The results show that RAG is most competitive on the longest inputs, where
retrieval can reduce the amount of irrelevant or truncated context presented
to the model. For shorter inputs, however, Direct inference generally
performs better, suggesting that retrieval can discard evidence required for
cross-document synthesis when the complete context can already be processed
directly.

These results help explain the domain-level pattern observed in
Section~\ref{subsec:system_variant}: retrieval is more useful when context
length itself becomes a substantial constraint, but less effective when the
task depends on evidence distributed across multiple documents or sections.

\section{Data Collection Details}
\label{app:data_collection}

We curated our dataset exclusively from publicly available, real-world long-form documents. All documents were initially collected in PDF format and then parsed into markdown format to facilitate processing.

\paragraph{Academic Domain}
We sourced documents from Semantic Scholar\footnote{https://www.semanticscholar.org/} and Papers With Code\footnote{https://paperswithcode.com/}. Semantic Scholar provides a vast repository of academic literature, offering rich metadata on citation contexts and types, which is crucial for tasks requiring analysis of inter-paper relationships. Papers With Code offers leaderboard-style compilations of results for various datasets, providing structured experimental outcomes reported in research papers. All documents collected for this domain were in English.

\paragraph{Financial Domain}
We collected A-share company annual reports and their summaries of both Chinese and English versions published between 2020 and 2023, sourced from CNINFO\footnote{https://static.cninfo.com.cn/}. These reports are comprehensive, lengthy, and contain a wealth of structured financial data and textual disclosures, ideal for complex table construction tasks. This domain specifically contributes bilingual (Chinese and English) data to AOE.

\paragraph{Legal Domain}
We selected Chinese civil law judgments due to their codified nature, unlike precedent-reliant common law systems. This structure is ideal for systematic retrieval, reasoning, and applying statutes. Our dataset draws from two authoritative sources: (1) The People's Court Case Library\footnote{https://rmfyalk.court.gov.cn} (providing reference and guiding civil cases) and (2) The China National Database of Laws and Regulations\footnote{https://flk.npc.gov.cn/index.html} (offering official statutory provisions).

\paragraph{Medical Domain}
We sourced biomedical case report papers from PubMed Central (PMC) Open Access Subset\footnote{https://pmc.ncbi.nlm.nih.gov/}. The data construction pipeline consists of three stages:

\textbf{Stage 1: Review Paper Discovery.}
We queried PubMed with the following search strategy to identify high-quality review articles containing structured case summaries:
\begin{quote}
\small
\texttt{("case reports"[Publication Type]) AND ("literature review"[Publication Type] OR "systematic review"[Publication Type]) OR ("table"[Text Word] OR "summary of cases"[Text Word]) AND ("open access"[Filter])}
\end{quote}
We restricted results to Q1/Q2 journals (by JCR ranking) to ensure data quality. From the retrieved reviews, we selected those containing ``Summary of reported cases'' tables---structured tables where each row corresponds to an independent case report paper and columns represent clinical attributes (e.g., age, diagnosis, treatment, outcome).

\textbf{Stage 2: Source Paper Collection and Filtering.}
For each row in a selected review table, we retrieved the original case report paper cited therein. Two filtering criteria were applied: (1) the cited paper must be Open Access (ensuring legal redistribution), and (2) each markdown document must describe exactly one clinical case (ensuring a clean one-row-per-document mapping). Papers failing either criterion were excluded, and the corresponding rows were removed from the ground-truth table.

\textbf{Stage 3: Human Annotation and Quality Control.}
Annotators verified each instance to ensure: (a) no duplicate rows exist, (b) column names are clinically meaningful and consistent across instances within the same review, (c) the number of target rows is manageable (3--15 per instance), and (d) cell values in the ground-truth table are traceable to specific content in the source papers. The review table itself serves as the gold standard, leveraging the domain expertise of the original review authors.

This pipeline yields ground-truth tables that are inherently authoritative---they are curated by domain experts (the review paper authors) and published in peer-reviewed journals, providing a stronger quality guarantee than crowd-sourced annotations.

\section{Detailed Annotation Process with Examples}
\label{sec:appendix_annotation}

To further illustrate the two-phase annotation process described in the main text, this section provides a concrete, step-by-step example from the Academic domain. This walkthrough demonstrates how we ensure the accuracy and consistency of the ground-truth data in the AOE benchmark.
\subsection{Annotator Information}

Our annotation process involved a multi-stage, expert-driven approach. The core annotation team consisted of graduate students from our institution with relevant domain expertise (e.g., in finance and law). In recognition of their significant intellectual contributions to the benchmark's creation—which included refining task definitions, developing annotation guidelines, and performing the initial data annotation and validation—they are credited as co-authors on this paper.

Additionally, to ensure the highest level of accuracy for specialized content, we recruited external domain experts (practitioners in the financial and legal fields) to conduct a final round of review and adjudication on complex cases. These experts were compensated for their time at a rate commensurate with their professional expertise.

\subsection{Example: Annotating Methodology Performance in the Academic Domain}

\begin{itemize}
    \item \textbf{Context:} The source documents are several computer vision research papers that report performance metrics on the MS COCO dataset.
    \item \textbf{Pharse 1: Expert-led Query Definition}

    A domain expert with a background in computer science formulates a precise, real-world query to guide the extraction process.
    
    \begin{tcolorbox}[colback=gray!5!white, colframe=gray!60!black, sharp corners, boxsep=2pt, top=2pt, bottom=2pt, left=3pt, right=3pt]
        \textbf{Defined Query:} "List the FID performance of the proposed methods in the paper on the MS COCO dataset."
    \end{tcolorbox}

    The expert then designs a structured table schema to hold the extracted information, defining the necessary columns and their expected content. This schema is then provided to the annotators.

    \begin{tcolorbox}[colback=gray!5!white, colframe=gray!60!black, sharp corners, boxsep=2pt, top=2pt, bottom=2pt, left=3pt, right=3pt]
    \textbf{table\_schema} \\
    \{``columns'': [ \\
    \quad \{``name'': ``paper\_name'', ``about'': ``Paper Title''\}, \\
    \quad \{``name'': ``method'', ``about'': ``Proposed Method''\}, \\
    \quad \{``name'': ``result'', ``about'': ``Performance on Specific Metrics''\}, \\
    \quad \{``name'': ``models\_and\_settings'', ``about'': ``Model Settings''\} \\
    ]\}
\end{tcolorbox}

    \item \textbf{Pharse 2: Dual Annotation and Final Adjudication}
    
    The source documents and the empty table schema are assigned to two separate annotators. Each annotator independently reads the papers to find the relevant values and populates the Table \ref{tab:adjudicated_table}. If the results were identical, the annotation was accepted. If there were discrepancies, the task was escalated to a senior reviewer for final judgment.


\end{itemize}

This structured, multi-stage process is applied across all domains in our benchmark to ensure the highest possible data quality.

\begin{table}[t]
    \centering
    \small

    \begin{tabularx}{\columnwidth}{@{}p{1.6cm}p{1.cm}p{0.7cm}X@{}}
        \toprule
        \textbf{paper\_name} & \textbf{method} & \textbf{result} & \textbf{settings} \\
        \midrule
        Paper A & GAN & 21.5 & Standard settings (Sec. 3) \\
        \addlinespace
        Paper B & StyleGAN & 4.40 & Style mixing, 1024$^2$, no trunc. \\
        \addlinespace
        Paper C & ADM & 2.97 & Classifier guide, 250 DDPM steps \\
        \bottomrule
    \end{tabularx}
    \caption{Final adjudicated ground-truth table after the annotation process.}
    \label{tab:adjudicated_table}
\end{table}

\section{Details of Tasks definition}
\label{sec:task_details}

Unlike conventional benchmarks where each document maps to a single output row, our benchmark employs a \textbf{dynamic row generation} strategy where the relationship between input documents ($M$) and output rows ($R_T$) is task-dependent and reflects real-world analytical requirements.

\subsection{Design Principles}

\paragraph{Inferential Reasoning Requirements} Tasks like $Legal_0$ (Figure~\ref{fig:legal_example}, up) demand semantic understanding beyond keyword matching. Models must interpret legal narratives and map them to precise statutory provisions through causal reasoning—a capability that distinguishes genuine comprehension from pattern recognition.

\paragraph{Atomic Granularity and Practical Value} By requiring extraction of structured, atomic data points rather than aggregated summaries, tasks like $Legal_1$ support high-value downstream applications (e.g., quantitative disparity analyses) while preventing degradation into conventional summarization.

Table~\ref{tab:tasks_definition} presents the complete query schemas and ground truth columns for all tasks, demonstrating how these principles manifest across domains.

\subsection{Row Generation Patterns Across Domains}

\paragraph{One-to-Many Mapping (Legal Domain).}
In tasks such as \textbf{$Legal_1$: Defendant Verdict Extraction} (Figure~\ref{fig:legal_example}, bottom), a single case document may yield multiple output rows when the case involves multiple defendants. Each row atomically captures verdict details (defendant name, charge, sentence term) for individual defendants, enabling downstream quantitative analyses such as sentencing disparity studies across demographic or jurisdictional factors.

\begin{figure}[t!]
    \centering

    \begin{minipage}{\linewidth}
        \centering
        \small
        \begin{tabular}{p{2.2cm} p{5cm}}
            \toprule
            \textbf{Input} & \textbf{Example Content} \\
            \midrule
            \textbf{Case Facts} & On July 4, 2013, Zou Moumou, Qi Mou, and Yu Moumou stole materials to manufacture drugs... \\
            \textbf{Legal Docs} & \textit{Full text of ``Criminal Law of the PRC''} \\
            \midrule
            \textbf{Output} & \texttt{["Criminal Law Article 23", "Article 347"]} \\
            \bottomrule
        \end{tabular}

        \vspace{0.15cm}

        \begin{tcolorbox}[
            colback=red!5!white,
            colframe=red!75!black,
            fonttitle=\bfseries,
            title=Design Principle: Inferential Reasoning,
            boxsep=1pt,
            left=2pt,right=2pt,top=2pt,bottom=2pt
        ]
        \small
        Requires \textbf{deep legal reasoning} to map narrative facts to precise statutory provisions—not achievable through keyword matching.
        \end{tcolorbox}
    \end{minipage}

    \vspace{0.4cm}

    \begin{minipage}{\linewidth}
        \centering
        \small
        \begin{tabular}{p{2.3cm} p{1.2cm} p{1.5cm} p{1.2cm}}
            \toprule
            \textbf{Case} & \textbf{Defendant} & \textbf{Charge} & \textbf{Term} \\
            \midrule
            Guan Case & Guan M. & Embezzlement & 8 yrs \\
            \addlinespace
            Guan Case & Zhao M. & Embezzlement & 1.5 yrs \\
            \addlinespace
            Xu Case & Xu M. & Embezzlement & 3.75 yrs \\
            \bottomrule
        \end{tabular}

        \vspace{0.15cm}

        \begin{tcolorbox}[
            colback=blue!5!white,
            colframe=blue!75!black,
            fonttitle=\bfseries,
            title=Design Principle: Structured Atomicity,
            boxsep=1pt,
            left=2pt,right=2pt,top=2pt,bottom=2pt
        ]
        \small
        Extracts \textbf{granular, structured data} enabling quantitative analysis (e.g., sentencing studies) rather than summarization.
        \end{tcolorbox}
    \end{minipage}

    \caption{
        Legal domain task design principles. 
        \textbf{Top:} $Legal_0$ demonstrates \textit{inferential reasoning}—models must understand legal narratives and map them to precise statutory provisions through semantic understanding.
        \textbf{Bottom:} $Legal_1$ exemplifies \textit{atomic granularity}, where a single case document produces multiple rows when multiple defendants are involved, enabling quantitative analyses.
    }
    \label{fig:legal_example}
\end{figure}

\paragraph{Document-to-Citation Mapping (Academic Domain).}
For \textbf{$Aca_0$: Citation Context Extraction}, rows represent unique citation linkages between documents. A single paper citing $k$ references generates $k$ rows, each containing the citation context and metadata. This design supports bibliometric analyses and citation pattern studies.

\paragraph{Temporal Decomposition (Financial Domain).}
\textbf{Longitudinal financial tasks} ($Fin_0$--$Fin_3$) adopt a temporal row structure where each row corresponds to a specific reporting period. For instance, analyzing a company's 5-year performance from a single document containing multi-year data produces $R_T=5$ rows, with each row capturing period-specific metrics.

\paragraph{Cross-Entity Aggregation (Financial Domain).}
In \textbf{cross-company benchmarking tasks} ($Fin_4$--$Fin_6$), each row represents a distinct entity. Analyzing $M$ companies' annual reports yields $R_T=M$ rows, enabling comparative ratio analyses and peer benchmarking.

\section{RAG Analysis}
\label{app:rag_analysis}


\subsection{Direct vs.\ RAG}
We first compare RAG with Direct inference across Academic, Legal, and
Financial tasks. Table~\ref{tab:rag_comparison} reports results for five
representative models.

\begin{table*}[t]
\centering
\setlength{\tabcolsep}{4.0pt}
\renewcommand{\arraystretch}{1.08}
\small

\begin{tabular}{@{}ll
    rrr
    rrr
    rrr
    r@{}}
\toprule

\multirow{2}{*}{\textbf{Model}}
& \multirow{2}{*}{\textbf{Setting}}
& \multicolumn{3}{c}{\textbf{Academic}}
& \multicolumn{3}{c}{\textbf{Legal}}
& \multicolumn{3}{c}{\textbf{Financial}}
& \multirow{2}{*}{\makecell[c]{\textbf{Avg.}\\\textbf{Cell F1}}}
\\

\cmidrule(lr){3-5}
\cmidrule(lr){6-8}
\cmidrule(lr){9-11}

&
& \textbf{Col} & \textbf{Row} & \textbf{Cell}
& \textbf{Col} & \textbf{Row} & \textbf{Cell}
& \textbf{Col} & \textbf{Row} & \textbf{Cell}
&
\\
\midrule

\multirow{2}{*}{GPT-5.4}
& Direct
& 71.2 & 63.8 & \textbf{48.2}
& 72.0 & 71.0 & \textbf{40.1}
& 62.9 & 91.2 & \textbf{46.3}
& \textbf{44.9}
\\
& RAG
& 62.2 & 38.8 & 32.7
& 50.4 & 44.6 & 19.6
& 57.9 & 92.0 & 44.5
& 32.3
\\
\addlinespace[3pt]

\multirow{2}{*}{Claude Opus 4.6}
& Direct
& 71.2 & 57.1 & \textbf{36.3}
& 70.5 & 73.1 & \textbf{39.0}
& 60.9 & 90.3 & \textbf{46.7}
& \textbf{40.7}
\\
& RAG
& 67.9 & 37.1 & 23.1
& 51.8 & 60.6 & 28.6
& 57.0 & 95.7 & 45.0
& 32.2
\\
\addlinespace[3pt]

\multirow{2}{*}{Gemini 3.1 Pro}
& Direct
& 71.0 & 60.4 & \textbf{45.7}
& 69.8 & 76.3 & \textbf{49.7}
& 51.7 & 89.6 & 41.6
& \textbf{45.7}
\\
& RAG
& 59.1 & 46.8 & 34.2
& 54.5 & 62.7 & 22.1
& 50.4 & 100.0 & \textbf{45.9}
& 34.1
\\
\addlinespace[3pt]

\multirow{2}{*}{DeepSeek V4-Pro}
& Direct
& 71.2 & 77.5 & \textbf{58.6}
& 65.5 & 80.4 & \textbf{50.4}
& 51.8 & 87.5 & \textbf{39.9}
& \textbf{49.6}
\\
& RAG
& 59.2 & 43.6 & 28.5
& 41.9 & 69.8 & 31.0
& 48.2 & 74.8 & 34.2
& 31.2
\\
\addlinespace[3pt]

\multirow{2}{*}{Qwen 3.6-Plus}
& Direct
& 70.6 & 70.4 & \textbf{55.2}
& 65.9 & 84.4 & \textbf{55.2}
& 52.0 & 82.5 & 38.4
& \textbf{49.6}
\\
& RAG
& 67.8 & 46.1 & 33.3
& 49.7 & 68.3 & 31.9
& 51.7 & 95.4 & \textbf{41.4}
& 35.5
\\

\bottomrule
\end{tabular}

\caption{
\textbf{Direct vs.\ RAG performance across domains.}
Direct provides the full source documents, whereas RAG provides the top-$k$
retrieved chunks.
\textbf{Bold} indicates the higher Cell F1 between Direct and RAG for each
model--domain pair and for the overall average.
Col, Row, and Cell denote Column F1, Row F1, and Cell F1, respectively.
All values are reported in \%.
}
\label{tab:rag_comparison}
\end{table*}

RAG does not consistently improve structured extraction. Across Academic
and Legal tasks, RAG reduces Cell F1 for all five evaluated models, with
particularly large drops for DeepSeek-V4-Pro (58.6\% $\rightarrow$ 28.5\%
on Academic) and Gemini-3.1-Pro (49.7\% $\rightarrow$ 22.1\% on Legal).
The pattern differs on Financial tasks: RAG yields small improvements for
Gemini-3.1-Pro (41.6\% $\rightarrow$ 45.9\%) and Qwen-3.6-Plus
(38.4\% $\rightarrow$ 41.4\%), while slightly degrading the other models.

These results suggest that reducing the amount of input context through
retrieval does not by itself improve structured extraction. In particular,
retrieval can discard evidence required for tasks that depend on information
distributed across multiple passages. The relatively different behavior on
Financial tasks motivates a closer analysis of how RAG performance varies
with document length.

\subsection{Performance by Document Length}
\label{app:rag_length}

To better understand when retrieval-based generation becomes challenging, we construct a standardized evaluation subset of 140 cases, evenly split into \textbf{short} (70 cases) and \textbf{long} (70 cases) document groups. Table~\ref{tab:rag_context_results} reports the full results.

\begin{table}[t]
\centering
\small
\setlength{\tabcolsep}{3.5pt}
\renewcommand{\arraystretch}{1.08}
\begin{tabular}{llccc}
\toprule
\textbf{Model} & \textbf{Length}
& \textbf{Cell F1}
& \textbf{Col F1}
& \textbf{Row F1} \\
\midrule

\multirow{2}{*}{\textsc{Claude-Opus-4.6}}
& Short & 59.5 & 62.0 & 98.5 \\
& Long  & 41.8 & 51.7 & 92.9 \\
\midrule

\multirow{2}{*}{\textsc{DeepSeek-V3.2}}
& Short & 54.9 & 57.4 & 98.5 \\
& Long  & 37.8 & 48.9 & 88.0 \\
\midrule

\multirow{2}{*}{\textsc{Gemini-3.1-Pro}}
& Short & 55.0 & 53.2 & 100.0 \\
& Long  & 43.8 & 46.6 & 100.0 \\
\midrule

\multirow{2}{*}{\textsc{GPT-5.4-0305}}
& Short & 58.5 & 62.9 & 94.6 \\
& Long  & 41.6 & 52.6 & 89.3 \\
\midrule

\multirow{2}{*}{\textsc{Qwen-3.6-Plus}}
& Short & 54.6 & 56.3 & 100.0 \\
& Long  & 38.7 & 46.7 & 90.5 \\

\bottomrule
\end{tabular}
\caption{
\textbf{RAG performance by source-document length.}
We split the evaluation cases evenly into short and long source documents.
Across models, longer documents consistently reduce Cell F1 and Column F1,
while Row F1 is comparatively more stable.
All values are in \%.
}
\label{tab:rag_context_results}
\end{table}

Across all evaluated models, RAG performs substantially worse on the long
subset, with Cell F1 decreasing by 10--17 absolute percentage points.
Two factors may contribute to this degradation:
(1)~\textbf{increased retrieval noise}: as document length grows, the
retrieval index contains more chunks, making it harder for a fixed top-$k$
retrieval budget to consistently surface the relevant evidence; and
(2)~\textbf{fragmented evidence}: the evidence required for a single cell
may span multiple distant passages, while independent chunk-level retrieval
may fail to recover all relevant pieces within a single top-$k$ pass.
The following case study illustrates the first failure mode, where
semantically relevant retrieval fails to surface the factual evidence
required by the target table.

\subsection{Case Study: Semantic vs.\ Factual Retrieval}

Table~\ref{tab:case_study_rag_example} illustrates a representative failure. When queried for ``shareholder value creation analysis,'' RAG retrieves semantically related but factually useless passages (e.g., board statements about ``commitment to shareholder value''), while missing the numerical evidence (earnings per share, net assets) that is required to populate the target table. This example illustrates a limitation: standard embedding-based retrieval optimizes for topical similarity, whereas structured extraction demands precise factual recall---a gap that widens with document length.

\begin{table}[t!]
    \centering
    \small

    \begin{tabularx}{\columnwidth}{
        >{\raggedright\arraybackslash}X
        >{\raggedright\arraybackslash}X
    }
        \multicolumn{2}{c}{\textbf{Query:} Assess ``shareholder value creation analysis''} \\
        \midrule

        \cellcolor{green!15}\textbf{Required Evidence} &
        \cellcolor{red!15}\textbf{Context Retrieved by RAG} \\
        \midrule

        \texttt{Earnings per share (basic): \$1.15; Net assets per share: \$12.50; Total equity...}
        &
        \texttt{"Our board of directors remains fully committed to creating long-term shareholder value through strategic investments..."} \\
        \bottomrule
    \end{tabularx}

\caption{A RAG failure case. RAG retrieves semantically related passages (topical relevance) but misses the numerical evidence (factual precision) required for structured extraction.}
\label{tab:case_study_rag_example}
\end{table}

\section{Content Evaluation Details}
\label{app:cell_rating}

We provide additional details of the content evaluation procedure described
in Section~\ref{sec:cell_f1}. The overall procedure is summarized in
Algorithm~\ref{alg:row_matching}. Content evaluation consists of two stages:
cell-level semantic matching and table-level F1 aggregation.

\paragraph{Cell-level Semantic Matching.}
For each aligned predicted--gold cell pair, an LLM evaluator assigns a
semantic matching score in $[0,1]$. This soft matching is necessary because
a cell may contain complex or multi-item content for which exact string
matching would be overly restrictive.

For list or array values, the score reflects item-level coverage with an
additional penalty for incomplete or extraneous content. Specifically, we
apply a factor of $0.8$ to partial list matches. For example, given a gold
cell
\texttt{["Art.14", "Art.39", "Art.40"]}
and a prediction
\texttt{["Art.14", "Art.45"]},
one of the three gold items is correctly recovered, resulting in a score of
$(1/3)\times 0.8 = 0.27$.
Detailed evaluator prompts and scoring criteria are provided in
Appendix~\ref{app:prompt_details}.

\paragraph{Table-level F1 Aggregation.}
After cell-level scores are fixed, we aggregate them at the table level.
The true-positive contribution is the sum of semantic matching scores over
correctly aligned cells. Predicted cells absent from the ground-truth table
are counted as false positives, while gold cells missing from the prediction
are counted as false negatives. Cell precision, recall, and F1 are then
computed using the standard F1 formulation.

This two-level design separates semantic evaluation from table-level
aggregation: the LLM evaluator is used only to assess semantic similarity
within individual cells, while the final Cell F1 is computed
deterministically.

\section{Stability and Variance Analysis}
\label{app:multi_run}

To evaluate the robustness of our benchmark, we conduct a multi-run analysis: five representative LLMs (Claude-Opus-4.6, DeepSeek-V3.2, GPT-5.4, Kimi-K2.5, Qwen-3.6-Plus) are each evaluated across 5 independent runs. All metrics use macro-averaging (per-instance scores averaged over all instances; CSV-failed instances receive Cell F1 = 0).

\subsection{Run-to-Run Score Stability}

Table~\ref{tab:multi_run_stats} reports per-model, per-domain statistics across all four key metrics. The majority of model-domain combinations exhibit Cell F1 standard deviations below 3 percentage points, confirming that performance differences are statistically meaningful. Kimi-K2.5's Academic domain shows the highest variance (SD = 7.6\%), attributable to occasional CSV formatting failures that produce run-level score swings.

\begin{table*}[t]
    \centering
    \small
    \setlength{\tabcolsep}{3.5pt}
    \begin{tabular}{@{}ll rrrr@{}}
        \toprule
        \textbf{Model} & \textbf{Domain} & \textbf{Pass Rate} & \textbf{Col Recall} & \textbf{Row F1} & \textbf{Cell F1} \\
        \midrule
        \multirow{3}{*}{Claude-Opus-4.6}
        & Financial & 90.5 $\pm$ 6.0 & 69.3 $\pm$ 1.7 & 83.6 $\pm$ 4.0 & 45.9 $\pm$ 1.2 \\
        & Legal     & 93.2 $\pm$ 4.1 & 96.8 $\pm$ 1.1 & 72.4 $\pm$ 1.4 & 37.7 $\pm$ 1.8 \\
        & Academic  & 100.0 $\pm$ 0.0 & 100.0 $\pm$ 0.0 & 57.0 $\pm$ 0.8 & 39.1 $\pm$ 1.7 \\
        \midrule
        \multirow{3}{*}{DeepSeek-V3.2}
        & Financial & 90.0 $\pm$ 3.2 & 64.6 $\pm$ 1.8 & 85.2 $\pm$ 2.0 & 42.4 $\pm$ 1.0 \\
        & Legal     & 93.8 $\pm$ 2.8 & 91.8 $\pm$ 2.2 & 73.4 $\pm$ 2.3 & 43.6 $\pm$ 2.3 \\
        & Academic  & 80.2 $\pm$ 3.3 & 100.0 $\pm$ 0.0 & 45.9 $\pm$ 1.9 & 37.8 $\pm$ 1.2 \\
        \midrule
        \multirow{3}{*}{GPT-5.4}
        & Financial & 93.9 $\pm$ 3.3 & 68.9 $\pm$ 2.4 & 87.3 $\pm$ 3.8 & 44.6 $\pm$ 3.1 \\
        & Legal     & 93.1 $\pm$ 4.2 & 97.4 $\pm$ 0.6 & 65.5 $\pm$ 3.4 & 38.3 $\pm$ 2.7 \\
        & Academic  & 100.0 $\pm$ 0.0 & 99.6 $\pm$ 0.4 & 64.2 $\pm$ 1.1 & 47.7 $\pm$ 1.3 \\
        \midrule
        \multirow{3}{*}{Kimi-K2.5}
        & Financial & 93.8 $\pm$ 4.0 & 66.3 $\pm$ 2.0 & 80.1 $\pm$ 3.9 & 41.5 $\pm$ 4.4 \\
        & Legal     & 92.3 $\pm$ 3.2 & 95.4 $\pm$ 0.8 & 75.5 $\pm$ 6.3 & 43.1 $\pm$ 4.8 \\
        & Academic  & 97.2 $\pm$ 1.4 & 99.4 $\pm$ 0.7 & 62.9 $\pm$ 8.5 & 47.6 $\pm$ 7.6 \\
        \midrule
        \multirow{3}{*}{Qwen-3.6-Plus}
        & Financial & 93.8 $\pm$ 1.0 & 59.6 $\pm$ 2.6 & 83.1 $\pm$ 3.1 & 37.7 $\pm$ 2.3 \\
        & Legal     & 83.8 $\pm$ 4.7 & 91.4 $\pm$ 3.1 & 68.2 $\pm$ 5.8 & 41.7 $\pm$ 3.8 \\
        & Academic  & 96.6 $\pm$ 1.3 & 99.7 $\pm$ 0.6 & 66.9 $\pm$ 1.2 & 50.7 $\pm$ 1.1 \\
        \bottomrule
    \end{tabular}
    \caption{Multi-run performance (Mean $\pm$ SD) across 5 independent runs per model. CSV-failed instances score 0 and are included in the denominator. All values in \%.}
    \label{tab:multi_run_stats}
\end{table*}

\subsection{Ranking Stability}

Table~\ref{tab:ranking_stability} examines whether model \textit{rankings} are stable across runs---a stricter test than absolute score stability.

\begin{table}[h]
    \centering
    \small
    \begin{tabular}{@{}lcccccc@{}}
        \toprule
        \textbf{Model} & \textbf{R0} & \textbf{R1} & \textbf{R2} & \textbf{R3} & \textbf{R4} & \textbf{Range} \\
        \midrule
        Qwen-3.6-Plus & 1 & 3 & 3 & 2 & 2 & 1--3 \\
        Kimi-K2.5 & 3 & 1 & 1 & 4 & 1 & 1--4 \\
        GPT-5.4 & 2 & 2 & 2 & 1 & 3 & 1--3 \\
        DeepSeek-V3.2 & 4 & 5 & 4 & 3 & 5 & 3--5 \\
        Claude-Opus-4.6 & 5 & 4 & 5 & 5 & 4 & 4--5 \\
        \bottomrule
    \end{tabular}
    \caption{Overall rankings (macro-averaged Cell F1 across 3 domains) per run. Kendall's $W = 0.664$. The top tier (Qwen, Kimi, GPT) and bottom tier (DeepSeek, Claude) are distinguishable, but within-tier ordering fluctuates.}
    \label{tab:ranking_stability}
\end{table}

The overall concordance ($W = 0.664$) reflects moderate ranking stability. The top three models (Qwen-3.6-Plus, Kimi-K2.5, GPT-5.4) consistently outperform the bottom two (DeepSeek-V3.2, Claude-Opus-4.6), but within-tier ordering varies across runs due to tight score clustering (top-tier mean: 43--44\%, bottom-tier: 41\%). Per-domain concordance ranges from $W = 0.464$ (Financial) to $W = 0.712$ (Legal), indicating that Legal rankings are the most robust while Financial rankings are least stable.

\subsection{Value Type Stability}

Table~\ref{tab:value_type_stats} reports Cell F1 by answer value type, averaged across all 5 models and 5 runs.

\begin{table}[h]
    \centering
    \small
    \begin{tabular}{@{}lrr@{}}
        \toprule
        \textbf{Value Type} & \textbf{Mean Cell F1} & \textbf{SD} \\
        \midrule
        Numeric & 49.9 & 1.9 \\
        Calculation & 41.8 & 2.1 \\
        Reasoning & 29.6 & 3.4 \\
        Text & 28.8 & 1.6 \\
        \bottomrule
    \end{tabular}
    \caption{Cell F1 (\%) by answer value type, averaged across 5 models $\times$ 5 runs. Low SD confirms metric stability across value categories.}
    \label{tab:value_type_stats}
\end{table}

Numeric cells achieve the highest Cell F1 (49.9\%), as they involve constrained value lookup. Calculation cells score lower (41.8\%), reflecting multi-step arithmetic difficulty. Reasoning and text cells prove most challenging (${\sim}$29\%), requiring cross-document inference or open-ended synthesis respectively.

\section{Schema Guidance: Case Study}
\label{app:schema_guided}

\begin{table*}[t]
\centering
\small
\setlength{\tabcolsep}{3pt}
\renewcommand{\arraystretch}{1.05}

\resizebox{\textwidth}{!}{%
\begin{tabular}{@{}p{2.5cm} p{5.5cm} p{7.0cm} p{5.5cm}@{}}
\toprule
& \textbf{Gold Standard} & \textbf{Raw Prediction (Cell F1 = 0.0)} & \textbf{Schema-Guided (Cell F1 = 0.64)} \\
\midrule

\multicolumn{4}{@{}l}{\textit{Instance: fimmu-17-1806045 --- Takayasu Arteritis with IBD (5 rows, 5 target columns)}} \\
\midrule

Schema &
No\_references, Predominant\_phenotype, Number\_of\_cases, Treatment, Outcome &
\colorbox{red!15}{LVV\_type}, \colorbox{red!15}{IBD\_type}, \colorbox{red!15}{Age}, \colorbox{red!15}{Sex}, \colorbox{red!15}{Diagnostic\_order}, \colorbox{red!15}{Key\_presentation}, Treatment, Outcome &
No\_references, Predominant\_phenotype, Number\_of\_cases, Treatment, Outcome \\[4pt]

Col Recall &
--- &
\colorbox{red!15}{40\%} (2/5 gold cols) &
\colorbox{green!15}{100\%} (5/5 gold cols) \\[4pt]

Row 1 &
3 (56), TAK+CD, 2, anti-TNF$\alpha$, Improved &
{\cellcolor{red!15}TAK, CD, 16, M, Simultaneous, Abdominal pain, Prednisone, Asymptomatic} &
3 (56), TAK+CD, 2, anti-TNF$\alpha$, Improved \\[4pt]

Failure mode &
--- &
Adds 6 unrequested columns; disaggregates summary rows into per-patient entries &
Correct schema; minor value differences \\

\bottomrule
\end{tabular}%
}

\caption{
    Case study illustrating schema divergence without guidance. Without schema guidance (Prompt-), the model constructs a clinically reasonable but evaluation-incompatible schema: it invents 6 columns not in the gold standard and disaggregates summary rows into individual patient entries. Only 2 of 5 gold columns are recoverable after alignment, yielding Cell F1 = 0. With Schema-Guided prompting, the same model produces a perfect schema match (Column Recall = 100\%) and achieves Cell F1 = 0.64.
}
\label{tab:case_study_schema_guided}
\end{table*}

Table~\ref{tab:case_study_schema_guided} illustrates the dominant failure mode. Without schema guidance, Claude-Sonnet-4.6 produces a table that is \textit{clinically sound}---it extracts age, sex, disease subtype, and diagnostic sequ
ence, all factually correct. However, this ``ideal'' schema shares only 2 columns with the gold standard (Treatment, Outcome), making 60\% of extracted information invisible to evaluation. Worse, the model disaggregates the gold's summary-level rows (e.g., ``2 cases of TAK+CD'') into individual patient entries, causing row-level misalignment.

This case demonstrates why schema guidance is not merely an evaluation convenience but a \textit{design necessity}: without it, models and ground-truth operate in different semantic spaces, making Cell F1 an unreliable signal of extraction quality.

\section{Human Reference Study}
\label{app:human_reference}

To provide additional context for task difficulty, we conduct a small-scale
human reference study on 16 AOE instances spanning all four domains.
A graduate-level annotator follows the same schema guidance provided to the
models and is allowed to use online search tools during annotation,
approximating an open-book extraction setting.
The annotator does not have specialized expertise in all evaluated domains,
particularly the Academic and Medical tasks.
We therefore treat these results as a limited human reference rather than
a strict human performance ceiling.

\paragraph{Setup.}
The human annotations are evaluated using the same cell-level evaluation
pipeline as the model outputs.
Because the human study covers only a small subset of the benchmark, these
results are intended to contextualize domain-level task difficulty rather
than provide a controlled human--model comparison.

\begin{table}[t]
\centering
\footnotesize
\setlength{\tabcolsep}{2.8pt}
\renewcommand{\arraystretch}{1.05}

\begin{tabular}{@{}lcccc@{}}
\toprule
\textbf{System} 
& \textbf{Acad.} 
& \textbf{Legal} 
& \textbf{Med.} 
& \textbf{Fin.} \\
\midrule

\textbf{Human}
& 42.4 
& \textbf{85.8} 
& \textbf{34.0} 
& \textbf{92.9} \\

\midrule

GPT-5.4
& 48.2 & 40.1 & 14.6 & 46.0 \\

DeepSeek-V4-Pro
& \textbf{58.6} & \textbf{50.4} & 16.5 & 38.7 \\

Claude-Opus-4.6
& 36.3 & 39.0 & \textbf{21.5} & \textbf{46.6} \\

Gemini-3.1-Pro
& 45.4 & 47.9 & 12.9 & 40.5 \\

\bottomrule
\end{tabular}

\caption{
\textbf{Human reference and model performance across domains.}
Cell F1 (\%). 
}
\label{tab:human_performance}
\end{table}

\paragraph{Results and Analysis.}
The human reference reveals substantial differences in the sources of
difficulty across domains.
The annotator achieves 92.9 Cell F1 on Financial and 85.8 on Legal.
Financial tasks primarily require locating specific figures in very long
inputs, which can reach approximately 1.8M characters, while Legal tasks
require identifying masked statutes and separating rulings across multiple
parties within the same case.
The strong human performance suggests that, with sufficient time and access
to search tools, many instances in these domains are tractable for a careful
human, leaving substantial headroom for current models.

In contrast, the annotator achieves only 42.4 Cell F1 on Academic and 34.0
on Medical.
Academic tasks require reading research papers and extracting precise
experimental results, while Medical tasks require interpreting and
summarizing clinical information from case reports.
Given the annotator's limited expertise in these domains, these results
suggest that specialized domain knowledge is an important component of
their difficulty.

Taken together, these results help distinguish different sources of
difficulty in AOE. Financial and Legal tasks remain challenging for models
even when a tool-assisted human performs well, whereas Academic and Medical
tasks additionally place substantial demands on domain-specific
interpretation. This pattern is also consistent with our error decomposition
in Appendix~\ref{app:error_analysis}, where textual synthesis is among the
most challenging cell categories across models. The human results therefore
provide complementary evidence that benchmark difficulty reflects both
information-access challenges and domain-specific interpretation, rather
than a single source of failure.

\onecolumn     

\begin{algorithm}
\caption{Simplified Row Matching and Evaluation Algorithm}
\label{alg:row_matching}
\begin{algorithmic}[1]
\Procedure{CalculateRowMatching}{$P, G, key\_cols, llm\_prompt$}
    \State \textbf{Input:} Predicted table $P$, Gold table $G$, Key columns $key\_cols$, LLM prompt $llm\_prompt$
    \State \textbf{Output:} Precision $P_r$, Recall $R_r$, F1 score $F1_r$, Matched pairs
    \Statex
    \State // Build key mappings
    \State $P_{keys} \gets \{(i, BuildKey(P[i], key\_cols)) \mid i \in P\}$
    \State $G_{map} \gets \{key \to [indices] \mid key = BuildKey(G[j], key\_cols)\}$
    \State $matched \gets \emptyset$, $available_G \gets \{0,1,...,|G|-1\}$
    \Statex
    \State \Comment{1. Exact matching}
    \For{$(p_i, p_{key})$ in $P_{keys}$}
        \If{$p_{key} \in G_{map}$ and $G_{map}[p_{key}] \cap available_G \neq \emptyset$}
            \State $g_j \gets$ pick one from $G_{map}[p_{key}] \cap available_G$
            \State $matched \gets matched \cup \{(p_i, g_j)\}$
            \State $available_G \gets available_G \setminus \{g_j\}$
        \EndIf
    \EndFor
    \Statex
    \State \Comment{2. LLM semantic matching for unmatched rows} 
    \State $unmatched_P \gets \{(i,key) \mid (i,key) \in P_{keys}, i \notin matched.p\_indices\}$
    \State $candidate_{keys} \gets \{BuildKey(G[j], key\_cols) \mid j \in available_G\}$
    \For{$(p_i, p_{key})$ in $unmatched_P$}
        \State $best_{key} \gets CallLLM(p_{key}, candidate_{keys}, llm\_prompt)$
        \If{$best_{key} \neq "None"$ and $best_{key} \in G_{map}$}
            \State $g_j \gets$ pick one from $G_{map}[best_{key}] \cap available_G$
            \If{$g_j$ exists}
                \State $matched \gets matched \cup \{(p_i, g_j)\}$
                \State $available_G \gets available_G \setminus \{g_j\}$
            \EndIf
        \EndIf
    \EndFor
    \Statex
    \State \Comment{3. Calculate metrics}
    \State $N_m \gets |matched|$, $N_p \gets |P|$, $N_g \gets |G|$
    \State $P_r \gets N_m/N_p$, $R_r \gets N_m/N_g$
    \State $F1_r \gets 2P_r R_r/(P_r + R_r)$ (if $P_r + R_r > 0$)
    \State \Return $P_r, R_r, F1_r, matched$
\EndProcedure
\end{algorithmic}
\end{algorithm}

\twocolumn  





\section{Detailed Error Analysis}
\label{app:error_analysis}

\begin{table*}[t]
\centering
\small
\setlength{\tabcolsep}{6pt}
\renewcommand{\arraystretch}{1.08}

\begin{tabular}{@{}lccc@{\qquad}cccc@{}}
\toprule
& \multicolumn{3}{c}{\textbf{Structure \& Coverage}}
& \multicolumn{4}{c}{\textbf{Cell F1 by Content Type}} \\
\cmidrule(lr){2-4}
\cmidrule(lr){5-8}

\textbf{Model}
& \textbf{Schema}
& \textbf{Row}
& \textbf{Coverage}
& \textbf{Numeric}
& \textbf{Calculation}
& \textbf{Reasoning}
& \textbf{Text} \\
\midrule

\multicolumn{8}{@{}l}{\textit{Open-Weight Models}} \\[1pt]

DeepSeek-V3.2
& 67.6 & \textbf{89.8} & 50.6
& 90.0 & 43.9 & 41.6 & 25.7 \\

Qwen-3.6-Plus
& 63.3 & 80.2 & 44.9
& 93.8 & 34.4 & 45.1 & 21.9 \\

Kimi-K2.6
& 75.8 & 80.5 & 54.0
& 85.0 & 46.8 & 44.4 & 26.6 \\

GLM-5.1
& 64.5 & 86.1 & 49.0
& \textbf{96.2} & 41.9 & 32.3 & 19.4 \\

Kimi-K2.5
& 71.9 & 81.4 & 51.5
& 81.2 & 44.4 & \textbf{46.3} & 20.5 \\

DeepSeek-V4-Pro
& 62.2 & 86.4 & 45.1
& \textbf{96.2} & 37.7 & 43.2 & 18.7 \\

\midrule
\multicolumn{8}{@{}l}{\textit{Proprietary Models}} \\[1pt]

Gemini-3.1-Pro
& 58.4 & 88.1 & 45.3
& 90.0 & 38.7 & 42.5 & 15.1 \\

Claude-Sonnet-4.6
& \textbf{77.0} & 84.1 & \textbf{55.8}
& 86.2 & 45.7 & 27.1 & \textbf{30.2} \\

Claude-Opus-4.6
& 71.2 & 84.5 & 49.2
& 86.2 & 45.3 & 38.4 & 21.3 \\

Gemini-3-Flash
& 69.8 & 86.9 & 55.7
& 87.5 & \textbf{58.5} & 31.7 & 19.9 \\

GPT-5.4
& 73.8 & 88.6 & 54.0
& 91.2 & 45.1 & 36.2 & 19.2 \\

Seed-2.0-Pro
& 69.9 & 85.0 & 53.9
& 83.8 & 53.6 & 43.4 & 21.8 \\

\bottomrule
\end{tabular}
\caption{
\textbf{Error analysis across construction stages and content types.}
We decompose performance into construction success
(\textbf{Schema}, \textbf{Row}, and cell \textbf{Coverage})
and Cell F1 across four content types.
Models generally recover rows successfully, but coverage drops substantially.
Among answered cells, direct numeric extraction is considerably easier than
calculation, reasoning, and textual synthesis.
All values are percentages; best results in each column are \textbf{bolded}.
}
\label{tab:error_analysis_full}
\end{table*}

Table~\ref{tab:error_analysis_full} provides the full error decomposition
under the offline setting, together with the raw matched/gold counts behind
the aggregate statistics reported in the main text.
Across models, row recovery is consistently high (80.2--89.8\%), whereas
cell coverage is substantially lower (44.9--55.8\%).
This gap indicates that recovering the overall table structure does not
necessarily translate into successfully populating its cells.

The decomposition by cell type reveals a similarly consistent pattern.
Models perform best on direct numeric extraction (81.2--96.2\%), while
cells requiring additional computation or reasoning are more difficult:
financial calculation ranges from 34.4--58.5\% and legal reasoning from
27.1--46.3\%.
Textual summarization is the most challenging category, with Cell F1 ranging
from only 15.1--30.2\%.
These results suggest that the main difficulty lies not only in locating the
appropriate rows and columns, but also in grounding, transforming, and
synthesizing evidence into the correct cell values.

\paragraph{Patient-case misalignment.}
Beyond missing or incorrectly generated cells, we observe cases in which the
model retrieves plausible evidence but associates it with the wrong entity.
Table~\ref{tab:case_study_medical_full} shows a representative example from
the Medical domain under the schema-guided setting.
Claude-Opus-4.6 recovers all target columns and rows
(Col Recall $=1.0$, Row F1 $=1.0$), yet achieves only 0.61 Cell F1.

The model correctly extracts Row~2, but populates Rows~1 and~4 with evidence
from the same patient case (``Posterior clinoid'', age 34).
The resulting table is therefore structurally correct while several cell
values are assigned to the wrong patients.
Some cells in Row~1, such as Sex and treatment, happen to match the gold
values because the patients share these attributes, rather than because the
correct patient was identified.

This case further illustrates the distinction between \emph{structural
recovery} and \emph{content grounding}: correctly recovering the target
rows and columns does not guarantee that the extracted evidence is associated
with the correct entity.

\begin{table*}[t]
\centering
\footnotesize
\setlength{\tabcolsep}{5pt}
\renewcommand{\arraystretch}{1.12}

\begin{tabular}{@{}llp{5.0cm}p{5.0cm}@{}}
\toprule
\textbf{Case} & \textbf{Field} & \textbf{Gold} &
\textbf{Prediction (Claude-Opus-4.6)} \\
\midrule

\multirow{6}{*}{\textbf{Row 1}}
& Location    & Middle fossa
& \cellcolor{red!12} Posterior clinoid \\
& Age         & 41
& \cellcolor{red!12} 34 \\
& Sex         & M
& M \\
& Symptoms    & Visual disturbance, diplopia
& \cellcolor{red!12} Visual disturbances, headache \\
& Treatment   & PSR
& PSR \\
& Improvement & No
& \cellcolor{red!12} Yes \\
\midrule

\multirow{6}{*}{\textbf{Row 2}}
& Location    & Foramen magnum & Foramen magnum \\
& Age         & 73 & 73 \\
& Sex         & M & M \\
& Symptoms    & Quadriparesis, headache & Quadriparesis, headache \\
& Treatment   & TR & TR \\
& Improvement & Yes & Yes \\
\midrule

\multirow{6}{*}{\textbf{Row 4}}
& Location    & Convexity
& \cellcolor{red!12} Posterior clinoid \\
& Age         & 25
& \cellcolor{red!12} 34 \\
& Sex         & M
& M \\
& Symptoms    & Headache, dizziness, speech disorders
& \cellcolor{red!12} Visual disturbances, headache \\
& Treatment   & TR
& \cellcolor{red!12} PSR \\
& Improvement & N/A
& \cellcolor{red!12} Yes \\

\bottomrule
\end{tabular}

\caption{
\textbf{Patient-case misalignment despite correct table structure.}
Claude-Opus-4.6 recovers all target columns and rows
(Col Recall $=1.0$, Row F1 $=1.0$) but achieves only 0.61 Cell F1.
Row~2 is correctly extracted, while Rows~1 and~4 are both populated with
evidence from the same patient case (``Posterior clinoid'', age 34).
Incorrect predictions are highlighted in red.
}
\label{tab:case_study_medical_full}
\end{table*}

\section{Prompt Examples}

--We use the following prompt format. The full prompt specifications and usage examples are provided in our supplementary materials/code repository.

\subsection{Prompt Examples for Generating Tables}

\subsubsection{Prompt Example of Offline Setting for LLMs}
\lstset{
    backgroundcolor=\color[RGB]{245,245,244},
    breaklines=true,
    basicstyle=\ttfamily\small,
    columns=flexible,
    keepspaces=true
}

\begin{lstlisting}
Instruction:
    Role: Academic Citation Relationship Analyzer
    Objective: Extract and analyze citation relationships between academic papers from Markdown files.
    
    # Workflow Steps
    1. Document_Parsing:
       - Extract paper title from H1 header (e.g., "# Title")
       - Identify citation markers: "(Author et al., YYYY)" or "[N]"
    
    2. Table_Schema_Design:
       - Columns: ["Referencing_Paper_Title", "Cited_Paper_Title", 
                   "Citation_Context", "Citation_Purpose"]
    
    3. Citation_Linking:
       - Match citation markers to reference entries
       - Extract 2 sentences before/after citation
       - Classify purpose: ["background", "methodology", 
         "comparison", "data_source", "supporting_evidence"]
    
    4. Data_Structuring:
       - Output flat CSV table
       - Wrap fields with commas/quotes in double quotes
    
    Query: {query}
    Input_doc: {input_doc}
    
    Output Format:
    <schema>
    Schema:
      - name: Cited_Paper_Title
        valid_type: exact_match
      - name: Referencing_Paper_Title
        valid_type: exact_match
      - name: Citation_Context
        valid_type: fuzzy_match
      - name: Citation_Purpose
        valid_type: categorical
    </schema>
    
    ```csv
    Cited paper title,Referencing paper title,Citation context,Purpose
    "Paper A","Paper B","...context1...",background
    "Paper A","Paper B","...context2...",background
    "Paper B","Paper A","...context3...",background
    ```
\end{lstlisting}

\subsubsection{Prompt Example of Online Settings for Agents and DeepReasearch}
\lstset{
    backgroundcolor=\color[RGB]{245,245,244},
    breaklines=true,
    basicstyle=\ttfamily\small,
    columns=flexible,
    keepspaces=true
}

\begin{lstlisting}
 Instruction:
    You are a financial data analysis expert. Based on the given Query and Input_doc, extract specified information from financial statements and output it in CSV format.
    Please follow the steps below and output the results.

    **Step 1: Understand the task requirements and analyze the input data.**
    Since the Input_doc contains only a list of filenames, treat this as an online scenario. You must use the web search function to find and locate the content of these documents. Briefly describe your search strategy and the key sources found in the <search_summary> section.
    
    **Step 2: Design the output table structure.**
    Design a Schema based on the Query's requirements, defining the column names and data types to ensure a standardized output format.
    
    **Step 3: Implementation.**
    Based on the analyzed document content from your web search results, extract the content relevant to the Schema headers, strictly adhering to the format and unit requirements of the Schema, and output a single-level table in CSV format.
    
    **Step 4: Format check.**
    Enclose fields containing special characters like commas or newlines in double quotes. In specific cases, consider replacing commas with semicolons, spaces, or removing them entirely, provided it does not alter the original meaning of the data.

    **When processing financial data, please note:**
    - Convert monetary amounts to the standard unit (e.g., 4.5 million -> 4,500,000).
    - Keep the unit for share capital as "shares".
    - Extract the numerical value from percentages (e.g., 12.36% -> 12.36).
    - Convert negative values represented by parentheses to a negative number (e.g., "(1,234)" -> -1234).
    - Retain thousand separators in numerical values and enclose the field in double quotes (e.g., "1,234,567").
    - Round monetary values to the nearest whole number and ratios to two decimal places.
  Instruction_no_cot: |
    You are a financial data analysis expert. Based on the given Query and Input_doc, extract specified information from financial statements and output it in CSV format. If the Input_doc only contains filenames, use web search to find the document content.

  Output: |
    Query: {query}
    Input_doc: {input_doc}
    <search_summary>
     Briefly summarize your search strategy and the key document sources found here.
    </search_summary>
    Answers:
      <schema>
       The schema you built
      </schema>
      ```csv
       Your Output
      ```
    
  Example: |
    Example 1:
      Query: "For the specified documents, conduct a corporate profitability analysis."
      Input_doc:
        - "BBB_Group_2022_Q3_Report.pdf"
        - "BBB_Group_2022_Annual_Report.pdf"
      Answers:
        <search_summary>
         Performed a web search for "BBB Group 2022 Q3 Report" and "BBB Group 2022 Annual Report". Found the PDF versions of both reports on the official stock exchange's information disclosure website. Data was primarily extracted from the consolidated income statements.
        </search_summary>
        <schema>
          - name: Document Name
            about: Full filename of the financial report, string type.
          - name: Reporting Period
            about: The accounting period covered by the report (YYYY-Qn format).
            format: 2021-Q4
          - name: Total Operating Revenue ($)
            about: The value of the first line item in the income statement.
            valid_type: exact_match
          - name: Gross Margin (%)
            about: (Operating Revenue - Operating Cost) / Operating Revenue * 100, rounded to two decimal places.
          - name: Net Profit Attributable to Parent ($)
            about: Net profit attributable to the parent company from the consolidated income statement.
            valid_type: exact_match
          - name: Net Margin (%)
            about: Net Profit Attributable to Parent / Total Operating Revenue * 100, rounded to two decimal places.
          - name: Non-recurring P/L Ratio (%)
            about: Non-recurring profit or loss / absolute value of Net Profit * 100, rounded to two decimal places.
        </schema>
        ```csv
        Document Name,Reporting Period,Total Operating Revenue ($),Gross Margin (%),Net Profit Attributable to Parent ($),Net Margin (%),Non-recurring P/L Ratio (%)
        "BBB_Group_2022_Q3_Report.pdf",2022-Q3,"88,456,789,234",32.15,"9,345,678,123",
        10.56,2.34
        "BBB_Group_2022_Annual_Report.pdf",2022-Q4,"345,678,901,567",30.89,"38,901,234,567",
        11.25,1.07
        ```
\end{lstlisting}

\subsection{Prompt Examples for Evaluating Tables}
\label{app:prompt_details}
\subsubsection{Prompt Examples for LLM Overall Scores}

\lstset{
    backgroundcolor=\color[RGB]{245,245,244},
    breaklines=true,
    basicstyle=\ttfamily\small,
    columns=flexible,
    keepspaces=true
}

\begin{lstlisting}

Instruction: |
  # Evaluation Task Description
  As a professional evaluator, strictly compare the CSV output quality of Gold Answer and Predict Answer based on the Query. Evaluation dimensions include:

  ## Core Evaluation Dimensions

  1. Intent Understanding: <score1>{int}</score1>/100
      Evaluate the accuracy of Predict Answer's understanding of the data extraction or organization requirements in the Query. The core is whether the model grasps the Query's intent, identifies the types of data points to be extracted or organized, their relationships, and the core information the final table should reflect.
      * **Perfect Score (100):** Complete and accurate understanding of all Query requirements, with no deviations or omissions. Can clearly identify core tasks.
      * **High Score (80-99):** Accurate understanding of the main Query intent, but may have slight misunderstandings on minor details or implicit requirements that don't affect core task execution.
      * **Medium-Low Score (40-79):** Partial understanding bias or confusion about the core Query intent, capturing some information points but failing to fully distinguish different types of information.
      * **Low Score (1-39):** Serious understanding bias or fundamental misunderstanding of the core Query intent, directing the task in a completely different direction.
      * **Zero Score (0):** Output completely unrelated to Query requirements, or unable to determine any understanding of the Query.
      Score Range: 0-100

  2. Schema Construction: <score2>{int}</score2>/100
      Evaluate the completeness and accuracy of the Schema (column names/headers) constructed by Predict Answer. Whether the model correctly identifies and includes all necessary information fields required by the Query, excludes irrelevant fields, and whether field naming is clear and accurate.
      * **Perfect Score (100):** Constructed Schema completely matches Query requirements or Gold Answer structure, includes all necessary columns, with no redundant or incorrect columns.
      * **High Score (80-99):** Schema basically complete and correct, includes most necessary columns, may miss 1-2 minor columns.
      * **Medium-Low Score (40-79):** Schema has significant deficiencies, missing multiple important columns, or contains many irrelevant/incorrect columns.
      * **Low Score (1-39):** Schema extremely incomplete or error-ridden, missing almost all key columns.
      * **Zero Score (0):** Failed to generate any Schema (no headers), or generated Schema completely unrelated to Query requirements.
      Score Range: 0-100

  3. Content Accuracy: <score3>{int}</score3>/100
      Evaluate the correctness of **specific data content** filled in the table by Predict Answer. Based on the Schema constructed by Predict Answer, check whether the data values of each filled field are precisely consistent with the information in Input Doc or Gold Answer.
      * **Perfect Score (100):** All filled data values in the table accurately and error-free reflect the corresponding information in Input Doc.
      * **High Score (80-99):** Most data values are correct, with only minor errors (such as spelling errors, extra spaces, etc.).
      * **Medium-Low Score (40-79):** Many data values are inaccurate, or extracted data is related but not precise enough.
      * **Low Score (1-39):** Most filled data values are incorrect, irrelevant, or completely fabricated (hallucinations).
      * **Zero Score (0):** All data filled in the table is incorrect, irrelevant, or completely fabricated.
      Score Range: 0-100

  4. Format Compliance: <score4>{int}</score4>/100
      Evaluate whether the output file of Predict Answer strictly follows the format requirements specified in the Query or task description, especially CSV basic format (delimiters, quote usage, line breaks, file encoding, etc.).
      * **Perfect Score (100):** Output file strictly adheres to all format specifications, file can be read without errors by standard parsers.
      * **High Score (80-99):** Output file basically meets format requirements, can be read by standard parsers, but has minor format issues.
      * **Medium-Low Score (40-79):** Output file has format issues that cause partial parsing errors or warnings.
      * **Low Score (1-39):** Overall format of output file does not meet requirements (such as outputting Markdown, plain text, etc. instead of CSV format).
      * **Zero Score (0):** No output file generated, or output is completely incomprehensible gibberish.
      Score Range: 0-100

  ## Score Parsing Markers
  Final output must include parsable delimiters, and scores must be integers:
  - <score1>{int}</score1>  # Corresponds to Intent Understanding
  - <score2>{int}</score2>  # Corresponds to Schema Construction
  - <score3>{int}</score3>  # Corresponds to Content Accuracy
  - <score4>{int}</score4>  # Corresponds to Format Compliance

Output: |
    Query: {query}

    Gold Answer csv: {gold_csv_txt}
    Predict Answer csv: {predict_csv_txt}

    Output:
    <output>
      1. Intent Understanding: <score1>score1</score1>/100 | Reason:
      2. Schema Construction: <score2>score2</score2>/100 | Reason:
      3. Content Accuracy: <score3>score3</score3>/100 | Reason:
      4. Format Compliance: <score4>score4</score4>/100 | Reason:
    </output>

Examples: |
    Example 1 - Academic Citation Extraction:
      Query: "Identify possible citation relationships among the following articles, understand, and provide the specific content being cited from the cited article."

      Gold Answer csv:
        Cited paper title,Referencing paper title,Referenced content,Label
        More Than Spoken Words: Nonverbal Message Extraction and Generation,ConceptNet 5.5: An Open Multilingual Graph of General Knowledge,"Thus, we use the human-annotated argument pairs such as (someone holds an umbrella, it is raining) in the Chinese set of commonsense knowledge ConceptNet by regarding the two arguments as the input and the NM respectively.",methodology
        More Than Spoken Words: Nonverbal Message Extraction and Generation,Know What You Don't Know: Unanswerable Questions for SQuAD,"an extractor similar to the formulation and implementation to address the unanswerable questions in machine reading comprehension tasks or generator simply using a pre-defined answer such as empty or none as the NM of these instances",background

      Predict Answer csv:
        Referencing Paper Title,Cited Paper Title,Citation Context,Citation Purpose,Citation Marker
        More Than Spoken Words: Nonverbal Message Extraction And Generation,Breazeal et al. (2005),"Nonverbal messages (NM), such as facial expressions, body movements, and tones of voice, can complement or modify verbal messages as well as improve the teamwork efficiency and effectiveness of face-to-face communication.",background,(Breazeal et al. 2005)
        More Than Spoken Words: Nonverbal Message Extraction And Generation,Phutela (2015),"Nonverbal messages (NM), such as facial expressions, body movements, and tones of voice, can complement or modify verbal messages as well as improve the teamwork efficiency and effectiveness of face-to-face communication.",background,(Phutela 2015)
        More Than Spoken Words: Nonverbal Message Extraction And Generation,Csaky and Recski (2021),"NMs are also seldom included in existing text-based dialogue tasks that mainly focus on verbal messages.",background,(Csaky and Recski 2021)

      Output:
      <output>
        1. Intent Understanding: <score1>40</score1>/100 | Reason: Query requires "Identify possible citation relationships" (identify citation relationships), "understand" (understand), and "provide the specific content being cited from the cited article" (provide specific content from cited article). 
        2. Schema Construction: <score2>30</score2>/100 | Reason: The Schema constructed by the model (Referencing Paper Title, Cited Paper Title, Citation Context, Citation Purpose, Citation Marker) deviates from the core Query requirement (extract "referenced content"). 
        3. Content Accuracy: <score3>20</score3>/100 | Reason: Query requires providing accurate "specific content from the cited article". The "Citation Context" provided in the prediction is text from the citing paper, not precise content from the cited article, which is completely inaccurate in content type. 
        4. Format Compliance: <score4>95</score4>/100 | Reason: Output is standard CSV format, using commas as delimiters, data rows correspond to headers. Format compliance is very high.
      </output>

\end{lstlisting}

\subsubsection{Prompt of Column Mapping and Format checking from LLMs}
\label{app:mapping}

\lstset{
    backgroundcolor=\color[RGB]{245,245,244},
    breaklines=true,
    basicstyle=\ttfamily\small,
    columns=flexible,
    keepspaces=true
}

\begin{lstlisting}
mapping_predict_tables:
    Instruction: |
      step1:
        Analyze column name matching
        Find the correspondence between the Gold_column_names and the Predict_answers column names through analysis and semantic matching. This includes case-insensitive matching and matching aliases or synonyms.

        For example:
        'Reporting Period' is equivalent to 'Fiscal Year'.
        'Owner's Equity' is equivalent to 'Net Assets'.
        'Referenced Content' is equivalent to 'Citation Context'.
        'Corresponding Legal Provisions' is equivalent to 'Related Index'.
        'Label' is equivalent to 'Citation Purpose'.
        
        Hint: If a column name in Predict_answers is too general, it can be split based on the more specific column names found in Gold_column_names.

        The finally determined matching columns **intersection_list**, ensure that each item name in it comes from Gold_column_names.
      step2:
        Ensure that each item in the intersection_list comes from Gold_column_names.
      step3:
        Use the intersection_list as the CSV header names, and fill the corresponding content from Predict_answers into this CSV.
      step4:
        Format check, ensure the generated CSV can be processed by pd.read_csv. If numbers contain commas as thousand separators, this will cause CSV parsing issues. To ensure correct processing by `pd.read_csv`, these numbers with commas need to be enclosed in quotes.

    Output: |
      Input:
        Predict_answers:  
        ```csv
         {predict_txt}
        ```
        Gold_column_names:
         {gold_header_string}
        
      Output:
        I have already mapped predict_txt according to the gold_header_string column names. Below is the processed Predict_csv:
        ```csv
        [Complete corrected CSV]
        ```
\end{lstlisting}

\subsubsection{Prompt Examples for Mapping Evaluation (Cell F1)}
\label{app:cell_rating}

\lstset{
    backgroundcolor=\color[RGB]{245,245,244},
    breaklines=true,
    basicstyle=\ttfamily\small,
    columns=flexible,
    keepspaces=true
}

\begin{lstlisting}
Instruction: |
  You are an expert evaluator strictly comparing text extracted into a table cell based ONLY on the provided Gold Standard and Predicted Text for the given Column Name. Adhere ONLY to the rules below.
  
  Evaluation Rules:
    1. Exact/Semantic Match: If the Predicted Text perfectly matches the Gold Standard Text OR conveys exactly the same essential information (allowing for minor formatting differences like currency symbols/names, date formats 'YYYY-MM-DD' vs 'Month D, YYYY', synonyms for common words that do not change meaning, or ignoring harmless extra punctuation like a trailing comma), output a score of 1.0.
    
    2. Critical Error/Missing Key Info: If the Predicted Text contains factually incorrect information compared to the Gold Standard, is completely unrelated, OR misses crucial information present in the Gold Standard, output a score of 0.0.
    
    3. List/Array Content (if applicable): If the content represents a list of items (detectable by structure like ["a","b"], a, b, c, or line breaks):
        - Identify the distinct items in the Gold Standard list (count = N_gold).
        - Identify the distinct items in the Predicted list (count = N_pred).
        - Find the number of items present in both lists (count = N_correct). Treat items case-insensitively and ignore minor whitespace differences for matching list items.
        - Calculate Score = (N_correct / N_gold) * 0.8. If N_gold is 0, the score is 1.0 if N_pred is also 0, otherwise 0.0. Ensure score does not exceed 0.8.
        - Note: If Gold is ["A", "B"] and Predicted is ["A", "C"], N_correct=1, N_gold=2, score = (1/2)*0.8 = 0.4.
        - If the field is NOT list-like, apply rules 1, 2, or 4.
    
    4. Partial Match (Other Cases): If it's not a perfect match (Rule 1), not a critical error (Rule 2), and not a list (Rule 3), estimate the degree of semantic overlap and correctness. For example, if one minor detail is wrong but the rest is correct, assign a score between 0.5 and 0.9. If significant parts are correct but some key info is missing or wrong, assign 0.1 to 0.5. Use your judgment based on how much of the core information is preserved.
    
    5. Empty/Null Cases:
        - If Gold is empty/null/NA and Predicted is also empty/null/NA, score is 1.0.
        - If Gold is empty/null/NA but Predicted has content, score is 0.0.
        - If Gold has content but Predicted is empty/null/NA, score is 0.0.
  
Output: |
  Column Name: {column_name}
  Gold Standard: "{gold_text}"
  Predicted Text: "{predicted_text}"
  Match Score: <output>0|1|0.x</output>
  Note: No explanation needed, output directly as <output>0|1|0.x</output>

Examples: |
  Examples (Illustrative based on rules):
    # Rule 1 Example
    - Column: `Product Price`
      Gold: `"$19.99"`
      Predicted: `"19.99 USD"`
      Score: <output>1.0</output>

    # Rule 2 Example
    - Column: `Delivery Date`
      Gold: `"2023-05-15"`
      Predicted: `"May 16, 2023"`
      Score: <output>0.0</output>

    # Rule 3 Example (List)
    - Column: `Legal Articles`
      Gold: `["Article 14","Article 39","Article 40"]` # N_gold = 3
      Predicted: `Article 14, Article 45`              # N_pred = 2, N_correct = 1 ("Article 14")
      Score: <output>0.27</output>                     # (1/3) * 0.8 -> 0.27

    # Rule 3 Example (List)
    - Column: `Related References`
      Gold: `["Patent CN2019/001","DOI 10.1234"]` # N_gold = 2
      Predicted: `Patent CN2019/001`              # N_pred = 1, N_correct = 1
      Score: <output>0.4</output>                 # (1/2) * 0.8 = 0.4

    # Rule 4 Example (Tolerance)
    - Column: `Industry Ban Period`
      Gold: `Prohibited from transportation industry for 4 years`
      Predicted: `, Prohibited from transportation industry for four years`
      Score: <output>1.0</output>
    
    - Column: `Probation`
      Gold: `Probation for one year and six months`
      Predicted: `One year and six months`
      Score: <output>1.0</output>
    
    - Column: `Other Rulings`
      Gold: `None`
      Predicted: ``
      Score: <output>1.0</output>

    # Rule 4 Example (Partial)
    - Column: Description
      Gold: "Red widget model X-1, uses 2 AA batteries."
      Predicted: "Red widget version X-1, requires batteries." # Missing battery type/count
      Score: <output>0.7</output> # Subjective partial score

    # Rule 5 Example both empty
    - Column: Middle Name
      Gold: ""
      Predicted: ""
      Score: <output>1.0</output>
\end{lstlisting}
\section{Use of AI Assistants}
\label{app:ai_assistants}

LLMs were used as part of the experimental methodology, including as
evaluators for assessing model outputs under predefined evaluation criteria.
Separately, we used GitHub Copilot for code completion in routine programming
tasks and ChatGPT for language polishing and presentation-related assistance
during manuscript preparation. All research design, experimental decisions,
data annotation, result analysis and interpretation, and scientific
conclusions were conducted and verified by the authors.
\onecolumn

\begin{longtable}{@{} p{0.05\textwidth} p{0.21\textwidth} p{0.11\textwidth} p{0.13\textwidth} p{0.37\textwidth} @{}}

\toprule
\textbf{Task ID} & \textbf{Task Objective} & \textbf{Input Documents} & \textbf{Primary Keys} & \textbf{Output Columns} \\
\midrule
\endfirsthead 

\toprule
\textbf{Task ID} & \textbf{Task Objective} & \textbf{Input Documents} & \textbf{Primary Keys} & \textbf{Output Columns} \\
\midrule
\endhead 

\bottomrule
\multicolumn{5}{r@{}}{(Continued on next page)} \\
\endfoot 

\bottomrule
\caption{Details of Tasks Definition}
\label{tab:tasks_definition}
\endlastfoot 





$Aca_{\text{0}}$ & Identify potential citation relationships between scholarly articles and extract the specific text segment that is being referenced. & Multiple research papers. & \texttt{["Cited paper title", "Referencing paper title"]} & \texttt{["Referenced content", "Label"]} \\

$Aca_{\text{1}}$  & Enumerate the reported \{\{Metrics\}\} of proposed methodologies within a research paper, specifically on the \{\{Dataset\_Name\}\} dataset. & Multiple research papers. & \texttt{["paper\_name", "method", "result"]} & \texttt{["result", "models\_and\_settings"]} \\
\midrule

$Fin_{\text{0}}$ & Conduct an analysis of enterprise scale and growth trends. & Annual company reports from 2020-2023. & \texttt{["Filename"]} & \texttt{["Reporting Period", "Total Assets (CNY)", "Asset Growth Rate (\%)", "Revenue (CNY)", "Revenue Growth Rate (\%)", "Total stockholders' equity (CNY)", "Weighted Average Return on Equity (\%)"]} \\

$Fin_{\text{1}}$ & Perform a comprehensive assessment of profitability. & Annual company reports from 2020-2023. & \texttt{["Filename"]} & \texttt{["Reporting Period", "Revenue (CNY)", "Net Profit (CNY)", "Net Profit Margin (\%)", "Weighted Average Return on Equity (\%)", "Net Profit Growth Rate (\%)", "Net cash provided by operating activities (CNY)", "Net Cash Ratio (\%)"]} \\

$Fin_{\text{2}}$  & Execute a diagnostic analysis of financial risks. & Annual company reports from 2020-2023. & \texttt{["Filename"]} & \texttt{["Reporting Period", "Total Liabilities (CNY)", "Liability-to-Asset Ratio (\%)", "Operating Cash Flow Growth Rate (\%)", "Total Assets (CNY)"]} \\

$Fin_{\text{3}}$ & Analyze the creation of shareholder value. & Annual company reports from 2020-2023. & \texttt{["Filename"]} & \texttt{["Reporting Period", "Share Capital (Shares)", "Total stockholders' equity (CNY)", "Net Profit (CNY)", "Net Asset per Share (CNY/share)", "Net Profit per Share (CNY/share)", "Weighted Average Return on Equity (\%)"]} \\

$Fin_{\text{4}}$ & Benchmark the operating performance for the year \{\{YEAR\}\} across different companies within the same industry. & Annual reports from different companies in the same industry for the year \{\{YEAR\}\}. & \texttt{["Filename"]} & \texttt{["Reporting Period", "Revenue (CNY)", "Net Profit (CNY)", "Net cash provided by operating activities (CNY)"]} \\

$Fin_{\text{5}}$ & Analyze the balance sheet structure for the year \{\{YEAR\}\} across different companies. & Annual reports from different companies in the same industry for the year \{\{YEAR\}\}. & \texttt{["Filename"]} & \texttt{["Reporting Period", "Total Assets (CNY)", "Liabilities (CNY)", "Total stockholders' equity (CNY)", "Share Capital (Shares)", "Debt to Asset Ratio"]} \\

$Fin_{\text{6}}$ & Assess and benchmark the investment value for the year \{\{YEAR\}\} across different companies. & Annual reports from different companies in the same industry for the year \{\{YEAR\}\}. & \texttt{["Filename"]} & \texttt{["Net Profit (CNY)", "Total stockholders' equity (CNY)", "Weighted Average Return on Equity (\%)", "Earnings per Share (CNY)", "Equity per Share (CNY)"]} \\
\midrule

$Legal_{\text{0}}$ & Retrieve precise legal provisions relevant to given case descriptions. & Multiple case information documents and related legal statutes. & \texttt{["Case\_name"]} & \texttt{["Relevant\_index", "Case\_number", "Keywords", "Basic\_case\_facts"]} \\

$Legal_{\text{1}}$ & Extract the final verdict for each defendant from case information. & Multiple case information documents. & \texttt{["Case\_name", "Defendant"]} & \texttt{["Charge", "Sentence\_term", "Probation", "Fine", "Confiscation", "Deprivation", "Other\_judgments", "Case\_number", "Basic\_case\_facts"]} \\
\midrule

$Med_{\text{0}}$ & Extract structured clinical attributes from medical case reports. Schema derived from original table caption. & Multiple medical case report papers. & \texttt{["source\_doc"]} & Task-dependent (e.g., demographics, diagnosis, treatment, outcome, clinical measurements). Schema guidance provided. \\
\end{longtable}

\twocolumn

\end{document}